# Review of Disentanglement Approaches for Medical Applications
## Towards Solving the *Gordian Knot* of Generative Models in Healthcare

Jana Fragemann[1]   Lynton Ardizzone[2]   Jan Egger[1]   Jens Kleesiek[1]

**Abstract**

Deep neural networks are commonly used for medical purposes such as image generation, segmentation, or classification. Besides this, they are often criticized as black boxes as their decision process is often not human interpretable. Encouraging the latent representation of a generative model to be *disentangled* offers new perspectives of control and interpretability. Understanding the data generation process could help to create artificial medical data sets without violating patient privacy, synthesizing different data modalities, or discovering data generating characteristics. These characteristics might unravel novel relationships that can be related to genetic traits or patient outcomes. In this paper, we give a comprehensive overview of popular generative models, like Generative Adversarial Networks (GANs), Variational Autoencoders (VAEs), and Flow-based Models. Furthermore, we summarize the different notions of *disentanglement*, review approaches to disentangle latent space representations and metrics to evaluate the degree of disentanglement. After introducing the theoretical frameworks, we give an overview of recent medical applications and discuss the impact and importance of disentanglement approaches for medical applications.

## Introduction

Medical image analysis is a key concept for diagnosing diseases and creating or evaluating individual patients' treatment plans. For that purpose, deep neural networks are already used for image generation, classification, or segmentation tasks [Goodfellow et al., 2014; Simonyan and Zisserman, 2015; Ronneberger et al., 2015]. While being successful, there are still some drawbacks. Often a huge amount of labeled, clean and complete data is needed, prediction processes of neural networks are often difficult to follow, shortcut learning distorts predictions and generalization tasks are cumbersome, to only mention a few challenges. But deep learning models themselves can help to overcome these issues. With generative models, synthetic medical data sets could be created, offering access to faithful and task-specific data sets without violating patients' privacy and avoiding elaborated and error-prone manual data labeling. Furthermore, a human interpretable latent space representation gives rise to various opportunities. Next to automatically generating labeled data, it allows find-ing patterns in the data that can be used to characterize and quantify normal and pathological changes. In turn, these patterns can be related to genetic traits, patient outcomes and used to guide therapy. Furthermore, generative models can be used to deal with missing data and create different modalities from given scans. These predicted scans, here denoted as virtual images, could offer a time-efficient way to obtain more information about the patient's condition, without exposing the patient to further radiation and the stress of the image acquisition. While these challenges are partially addressed successfully by deep generative neural networks, a major drawback remains. The decision process is often intractable and, therefore, neural networks are often criticized as black boxes [Egger et al., 2021] or, as we envision it, as *Gordian Knot*, which could be solved exploring and disentangling the data representation hidden in the internal states of the neural networks. A key concept, therefore, is a human interpretable and manageable latent space representation of generative models. More technically, this is described as *disentanglement*. This term lacks a uniform definition, but, in general, it is about controlling the semantic changes in the data, with respect to their internal representation in the network. In recent years, there has been a surge of interest in this area of research. A highly recommended tutorial by Liu et al. [2021a] and a publication of Locatello et al. [2020] give a good overview. Nevertheless, this field is growing so fast and offers many possibilities, thus, we also give an overview and focus more on medical applications. In general, we would like to address the following questions:

1. What is a *disentangled* representation of a generative model?

2. Which methods exist to encourage a disentangled latent representation?

3. What is the added value for medical applications?

4. How to evaluate the degree of disentanglement?

5. What future medical applications can benefit from disentanglements?

We start with an introduction of our used notation and abbreviations. Then we introduce the ideas of the most popular generative models: Generative Adversarial Networks (GANs) [Goodfellow et al., 2014], Variational Autoencoders (VAEs) [Kingma and Welling, 2014] and Flow-based Models [Dinh et al., 2015]. Next, we give an overview of several theoretical definitions of disentanglement, consider different methods trying to achieve this, and give an overview of metrics to evaluate the degree of disentanglement. In the end, we systematically review medical applications of disentangled representations.

**Notation**

In the following tables we give an overview about the used notations and abbreviations in this paper. Equations are numbered only when they are referenced.

| Symbol | Description |
|---|---|
| $X, Z$ | random variables |
| $x, z$ | values of the random variables $X, Z$ |
| $x^{(i)}$, i = 1,...,N | $i$-th data point, where $N$ denotes the number of samples |
| $\mathcal{X}$ | set of all data points |
| $p(X)$ | distribution of the random variable $X$ |
| $p(x)$ | short notation for the evaluated density function of the distribution of the random variable $X$, $P(X = x)$ |
| $p_\theta, q_\phi$ | true distribution/model distribution, the index denotes the parameters describing the distribution, e.g. for a normal distribution $\theta$ would contain mean and variance |
| $\mathbb{E}_{z \sim q(Z)} p(z)$ | expected value of the density $p(z)$ where $z$ is sampled from $q(Z)$ |
| $\mathcal{Z}$ | latent space |
| $d$ | latent space dimension |
| $z_i$ | $i$-th dimension of the latent space vector |
| $K$ | ground truth space dimension |
| $g$ | ground truth vector |
| $enc, dec$ | encoder, decoder |
| $D, G$ | discriminator and generator network |

| Abbrevation | Meaning |
| --- | --- |
| CIFC | Cross Image Feature Consistency Error |
| CGAcc | Conditional Generation Accuracy |
| CNN | Convolutional Neural Network |
| CNR | Contrast-to-Noise-Ratio |
| DCI | Disentanglement, Informativeness, Completeness Metric |
| DNN | Deep Neural Network |
| ELBO | Evidence Lower Bound |
| ENL | Equivalent Number of Looks |
| FID | Frechet Interception Distance |
| GAN | Generative Adversarial Network |
| IS | Interception Score |
| *KL* | Kullback-Leibler-Divergence, see Appendix 7.2 |
| MIG | Mutual Information Gap |
| MTM | Manifold Topology Metric |
| PPL | Perceptual Path Length Metric |
| SAP | Separated Attribute Predictability Metric |
| SSIM | Structural Similarity Index |
| TC | Total Correlation |
| UDR | Unsupervised Disentanglement Ranking |
| VAE | Variational Autoencoder |

# 1 Generative Models

Machine learning methods can be divided into two types: *generative* and *discriminative* methods. Discriminative models yield good results for image classification or segmentation tasks [Ronneberger et al., 2015]. These models directly predict the probability that a given data point $x$ belongs to a class $y$, thus $p(y|x)$. Unlike standard classification or segmentation networks, the goal of generative models is not to produce some label or prediction from an input $x$. Instead, they learn to model the distribution of the inputs themselves, $p(X)$. The approximated distribution $q(X) \approx p(X)$ can then be used to create new data samples, perform generative classification, outlier detection, and more. To be able to model such a complex and high-dimension distribution, almost all generative models make use of the so-called reparametrization trick. This means, instead of parameterizing $q(X)$ directly, it is expressed in terms of a parameterized transformation $f_\theta$ between $q(X)$ and the latent space $\mathcal{Z}$, which has a known latent distribution $p(Z)$ prescribed. The goal of training is then to find such a transformation, represented by the network, that $q(X)$ that is obtained from transforming the latent distribution is as close as possible to the true distribution $p(X)$. In this context the following nomenclature is usually applied:

- *p(Z)*→ **prior distribution:**
  If we do not know anything about the data we want to know how likely is the unobserved variable value $z$? Hence, it describes the distribution of the latent space $\mathcal{Z}$ without consideration of the data, thus before knowing anything about the data $\mathcal{X}$.

- *p(X|Z)*→ **likelihood distribution:**
  The likelihood $p(X|Z)$ describes the distribution of $X$ if we already know $Z$. This means how likely is a value $x$ if we assume $Z$ to have a specific value?

- *p(Z|X)* → **posterior distribution:**
  The posterior $p(Z|X)$ is the distribution of $Z$, if we know $X$.

- *p(X)* → **evidence distribution:**
  $p(X)$ is the data distribution, which is denoted as evidence.

In general, three types of generative models are used in practice:

- Generative Adversarial Networks (GANs);
- Variational Autoencoder (VAEs);
- Flow-based Models.

Their latent code can also be used for downstream tasks in the context of representation learning or transfer learning.

## 1.1 Generative Adversarial Network

A Generative Adversarial Network (GAN) consists of two neural networks, a generator, and a discriminator, see Figure 1. Given an input image the discriminator should return the probability that this input is a sample from the true data distribution or not. This is done by maximizing:

$$\max_D \; \log(D(x)) + \log(1 - D(\hat{x}))$$

where $x$ are real data and $\hat{x}$ is generated by the model. Thus, by maximizing this term real data $x$ gets a high probability $D(x)$ and generated data $\hat{x}$ get a low probability $D(\hat{x})$. This enables us for unseen data to decide whether they are from the same distribution or not. But we want to be able to generate new samples from this distribution, not only decide if existing samples are probably drawn from the same distribution or not. Therefore, the second network, the generator $G$, is needed. This network generates the data $G(z) = \hat{x}$ from a latent space input $z \in \mathcal{Z}$ and tries to fool the discriminator network. Thus, it learns to generate data points $\hat{x}$ which have a high probability to be a sample from the data distribution. Therefore, $G$ learns to generate samples from a distribution $q(X)$ which approximates the true data distribution $p(X)$. This is done by minimizing:

$$\min_G \; \log(1 - D(G(z)))$$

As the name Generative *Adversarial* Network implies, training these both networks together results in an adversarial optimization process, also denoted as minmax game, which includes the expected value about all training data points:

$$V(D,G) = \min_G \max_D \mathbb{E}_{x \sim p(x)}[\log(D(x))]$$
$$+ \mathbb{E}_{z \sim p(z)}[\log(1 - D(G(z)))]$$

This minmax game is equivalent to minimize the Jensen-Shannon-Divergence of the real data distribution $p(X)$ and their approximation from the generator network $q(X)$, see Appendix 7.3. While these GAN approaches receive really good image quality for the generation of new samples, they suffer from training instabilities. This can be avoided by replacing the Jensen-Shannon-Divergence with the Wasserstein metric [Arjovsky et al., 2017].

An often applied extension of the GAN network is the CycleGAN [Zhu et al., 2017]. This network is used for a specific kind of image generation, known as image-to-image translation, e.g. translation of a *summer* landscape to its *winter* version. Therefore, CycleGAN consists of two generators and two discriminators.

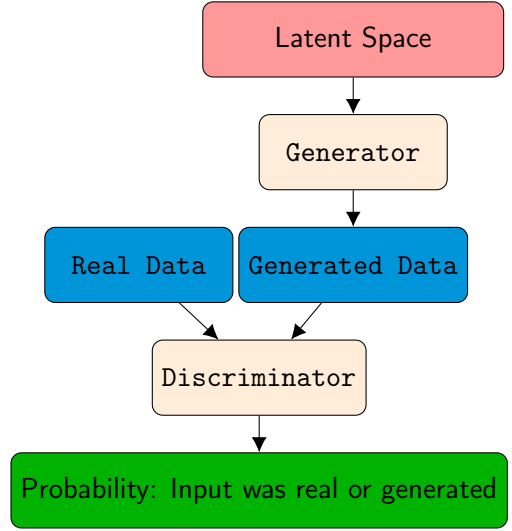

*Figure 1: Generative Adversarial Network*

## 1.2 Variational Autoencoder

Like Generative Adversarial Networks, Variational Autoencoders consist of two jointly trained neural networks, the encoder, and the decoder. The encoder maps the input data $x^{(i)}$ onto a latent space representation, which is taken to be the expected value of the posterior distribution $p(Z|X = x^{(i)})$. A sample from this distribution is the input for the decoder which returns the expected value of the likelihood distribution $p(X|Z = z)$. A combination of these two density functions describing these distributions builds a lower bound of the value of the density function of the evidence distribution:

$$\log(p(x^{(i)})) \geq \mathbb{E}_{z \sim q_\phi(Z|x^{(i)})}\big(\log(p_\theta(x^{(i)}|z))\big) \\ - KL\big(q_\phi(Z|x^{(i)})||p(Z)\big) \quad (1)$$

This bound is known as ELBO (Evidence Lower Bound). The first term on the right-hand side is often referred to as a kind of a reconstruction error and the second term is the Kullback-Leibler-Divergence, a measure for the difference of the two distributions. The derivation of (1) and the definition of the Kullback-Leibler-Divergence can be found in the Appendix 7.2.

This bound is maximized for every data point with respect to the distribution describing parameters $\theta$ and $\phi$. Therefore, the resulting loss function $L(\theta, \phi)$ is de-

fined as:

$$L(\theta,\phi) := \sum_{i=1}^{N} \mathbb{E}_{z \sim q_\phi(Z|x^{(i)})}\big(-\log(p_\theta(x^{(i)}|z))\big) \quad (2)$$
$$+ KL\big(q_\phi(Z|x^{(i)}) || p(Z)\big)$$

Thereby, to sample from the distribution $p(Z|x)$ Kingma and Welling [2014] introduced a *reparameterization trick*. For the univariate Gaussian case, a sample $z'$ is taken as:

$$z' = z + \sigma\epsilon$$

where $\epsilon \sim \mathcal{N}(0,1)$, $\sigma$ is the variance of $p(Z|x)$ and $z$ the output of the encoder.

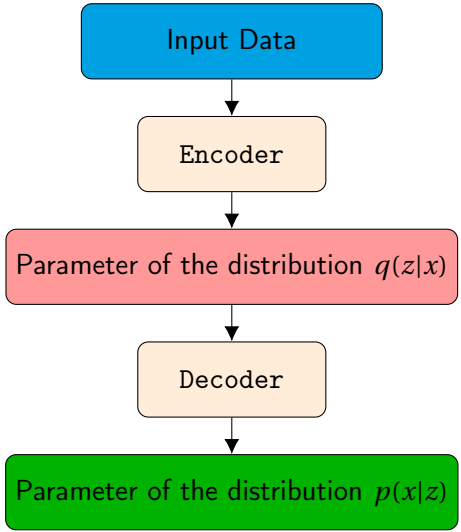

*Figure 2: Variational Autoencoder consist of two neural networks, the encoder and the decoder which learn the parameters of the posterior and likelihood distribution.*

## 1.3 Normalizing Flows

Normalizing flows follow a similar approach as VAEs: They learn a bi-directional mapping between data space and latent space in order to parameterize the data distribution $p_\theta(X)$. The main difference is that the network $f$ representing this transformation is exactly invertible (a diffeomorphism to be precise), and has a tractable Jacobian determinant. Therefore, the encoder and decoder can be represented by the same network, see Figure 3, which can simply be inverted depending on whether decoding or encoding is required. Separate complementary encoder and decoder networks are no longer needed. This has both profound theoretical and practical implications.

The effective differences come from the fact that the invertibility and the Jacobian determinant avoid the problem of $p_\theta(X)$ being intractable to compute exactly. This greatly simplifies the loss function and training procedure, and avoids the artifacts associated with VAEs, as explained below. To provide some necessary mathematical background, the Jacobian matrix $J$ is a matrix containing the partial derivatives of all network output dimensions with respect to all input dimensions:

$$J_{ij} = \frac{\partial f_i}{\partial x_j} \quad (3)$$

For a bijective function $f$, the number of input dimensions will be equal to the number of output dimensions, so $J$ will be a square matrix. While computing this matrix for standard neural networks is prohibitively expensive, most existing invertible network architectures provide the determinant $\det J$ for a low computational cost or for free, due to specialized architectures. Using the Jacobian determinant, the density $p_\theta(x)$ expressed by the model can be computed from the prescribed latent distribution $p(Z)$, generally a standard normal, by simple application of the change-of-variables formula:

$$p_\theta(x) = p\big(z = f_\theta(x)\big)|\det J(x)| \quad (4)$$

With the model's probability being able to be computed exactly, it is also possible to perform exact maximum likelihood training, which cannot be done with other generative models:

$$\mathcal{L}(\theta) := \sum_i -\log p_\theta(x^{(i)}) \quad (5)$$

The maximum likelihood loss has been used and studied extensively for many classical methods and can be shown to match the model's density $p_\theta(x)$ to the true data distribution $\hat{p}(x)$. To generate new samples at test time, random latent vectors $z^*$ can be sampled from the latent distribution, and be transformed back to the data space by inverting the network: $x_{\text{generated}} = f_\theta^{-1}(z^*)$.

Comparing the objective in Equation (5) to the corresponding VAE loss in Equation (2), it is clear that the training procedure of normalizing flows is significantly simpler. It does not require the ELBO approximation, there is no reconstruction loss term – perfect reconstruction is guaranteed at all times due to invertibility. This is not only conceptually more elegant and direct, but also has practical advantages, such as avoiding blurry generated images seen in VAEs due to the reconstruction part of the loss, or the mode-mixing artifacts that stem from the Gaussian posteriors in standard VAEs (explained in greater theoretical detail e.g. by Bousquet et al. [2017]). Compared to GANs, normalizing flows do not suffer from the typical problem of mode collapse: with the maximum likelihood loss, the model is penalized strictly for missing even small modes of the training data distribution, an important property for medical and scientific applications, where unlikely or rare cases are often the ones of interest. Furthermore, normalizing flows allow likelihood estimation in addition to generation, and do not require careful tuning and balancing of the generator and discriminator. Of course, the advantages described above also come at a cost, which is mostly of technical nature: invertible neural networks are more restricted in their architecture, generally have more parameters, and require longer to train. The most common invertible architectures used today are Real-NVP [Dinh et al., 2016] or GLOW [Kingma and Dhariwal, 2018], both using so-called affine coupling blocks which repeatedly perform affine transformations on subsets of the variables; the i-ResNet [Behrmann et al., 2019], which is regularized ResNet that has to be inverted at test-time using a numerical procedure; and FFJORD [Grathwohl et al., 2018], which is a specialized version of neural ODEs [Chen et al., 2018a] for normalizing flows.

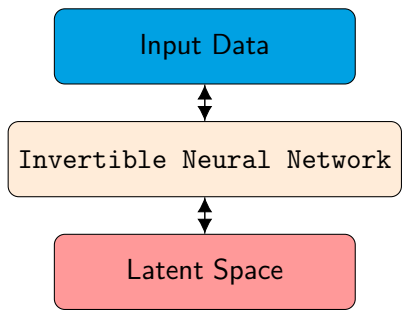

*Figure 3: Flow-based Models are invertible neural networks.*

## 2 Definition of Disentanglement

In the following, we will concentrate on image data for simplicity, but several other data modalities can be considered.

Assuming an image can be generated by a set of semantic meaningful features, like color, objects, shapes, etc. If these factors of variations would be captured in the latent space representation in a separate, thus disentangled, and interpretable way, the image generation process becomes understandable and controllable. The factors of variations are often denoted as generating factors and their numerical realizations are denoted as ground truth factors. E.g. in the dSprites data set [Matthey et al., 2017], which contains synthetic images of simple shaped objects, see Section 4.2, a factor of variation is the orientation of an object. Its numerical realization, thus the ground truth factor, is a value in the range of $[0, 2\pi]$. But how to check if a representation fulfills this property? Manually observing the change of the output of a generative model, while changing parts of the latent space representation is ineffective and subjective. Thus, before evaluating, if a representation is disentangled, a technical characterization has to be introduced. Different researchers have attempted to formalize the idea of a disentangled representation. Current investigations on a characterization are based on the comparison of the ground truth factors and the latent space representation, see Figure 4. As there is no agreement upon a definition, we shortly summarize three approaches to characterize a disentangled representation, even if several more exist [Higgins et al., 2018].

**Disentanglement, Completeness, Informativeness**

[Eastwood and Williams, 2018] and [Ridgeway and Mozer, 2018] simultaneously and independently introduced three characteristics of disentangled representations, denoted as *disentanglement, completeness* and *informativeness* [1]:

1. *Disentanglement*: One factor of the latent representation is only influenced by a change in one generative factor.

---
[1] Ridgeway and Mozer [2018] denoted these characteristics as *modularity, compactness* and *explicitness*.

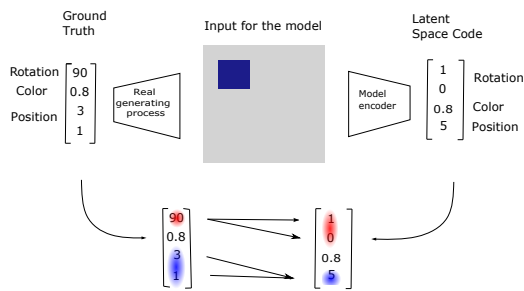

*Figure 4: Assuming that the image is generated from ground truth factors by a generating process. This image is an input for the deep learning model, which allocates a latent representation to it. Comparing the ground truth realization and the latent code on the bottom: In the ground truth, the rotation is parameterized by the angle (marked red), thus with one dimension of the latent space. In the latent space code, rotation is represented with the $sin$ and $cos$ values (marked red), thus, with two dimensions of the latent space. This example is taken from [Ridgeway and Mozer, 2018].*

2. *Completeness*: One generative factor is only sensitive to a change in one latent code dimension.

3. *Informativeness*: The amount of information captured by the latent code representation about the factors of variation.

Also Sepliarskaia et al. [2021] denote these three properties as the characteristics to define a disentangled representation. Thereby, the completeness characteristic is sometimes assumed to be not as important as the other ones, see Section 6.

**Informativeness, Separability, Interpretability**

The authors Do and Tran [2021] suggest a slightly different notion of disentanglement, based on information theory, with three criteria:

1. *Informativeness*: The mutual information between the data $x$ and a latent code $z_i$ should be high, where $z_i$ could contain more than one dimension.

2. *Separability and Independence*: The multivariate mutual information between data $x$ and two factors $z_i$ and $z_j$ should be zero. This means that $z_i$ and $z_j$ do not contain any redundant information about $x$.

3. *Interpretability*: The entropy of $z_i$, the entropy of $g_k$, which denotes a ground truth factor, and the mutual information between $z_i$ and $g_k$ should be the same.

**Consistency and Restrictiveness**

The authors of [Shu et al., 2020] describe the properties a disentangled representation should fulfill as:

1. *Consistency:* When one latent factor is fixed, the corresponding factor of variation does not change in the output.

2. *Restrictiveness:* While changing one latent factor and leaving all others fixed, only the corresponding factor of variation changes in the output.

All these definitions are characterizations of completely disentangled representations, with respect to certain known ground truth factors. Furthermore, these ground truth factors are assumed to be independent of each other. In practice, underlying factors of variation are often correlated and unknown. And even for known ground truth factors, the realizations can be different [Duan et al., 2019], e.g. the factor of variation describing the *rotation* can be represented with different numerical values and dimensions, see Figure 4. This is problematic for evaluating the degree of disentanglement, even if ground truth factors are known, see Section 4.

# 3 Disentanglement Approaches

Considering the latent space representations of GAN, VAE, or Flow-based methods, there exist model-agnostic and model-dependent attempts to encourage the latent space representation to be disentangled:

1. Training the model as usual and change the resulting latent space representation [Esser et al., 2020; Ren et al., 2021].

2. Modify the model loss function, with a:

- Model-agnostic approach [Peebles et al., 2020];
- Model-dependent modification [Higgins et al., 2017; Kim and Mnih, 2019; Chen et al., 2016].

As sometimes only specific factors want to be disentangled, another popular strategy is:

3. Train two networks simultaneously, one to learn one part of the factors of interest and another one to learn the remaining part [Ben-Cohen et al., 2019], which often needs some kind of supervision. Furthermore, the basic ideas to achieve a disentangled representation vary by the network's architecture, as we will see in the following section.

### 3.1 Unsupervised Models

The best known disentanglement models are $\beta$-*VAE* [Higgins et al., 2017] and *InfoGAN* [Chen et al., 2016], but several more exists. We divide the reviewed models into VAE-based, GAN-based, Flow-based, and model agnostic disentanglement approaches. An overview of all models and a timeline of the publications can be found in Figure 5 and Table 1.

#### 3.1.1 VAE-based Approaches

**$\beta$-VAE**

The $\beta$-*VAE*, introduced by [Higgins et al., 2017], is a Variational Autoencoder whose loss function is slightly modified with a parameter $\beta > 1$. This parameter gives more weight to the Kullback-Leibler-Divergence term in the original ELBO, see Equation (1), and therefore, the new loss function is defined as:

$$L(\theta, \phi) := \sum_{i=1}^{N} \mathbb{E}_{z \sim q_\phi(Z|x^{(i)})}\big(-\log(p_\theta(x^{(i)}|z))\big)$$
$$+ \beta KL\big(q_\phi(Z|x^{(i)})||p(Z)\big)$$

As the prior distribution $p(Z)$ is usually assumed to be a factorized standard normal distribution, thus, $p(Z) = \prod_{i=1}^{d} p(Z_i)$, where all $p(Z_i)$ are standard normal distributions of the $i-th$ dimension of a latent space representation. This modification should encourage the latent space factors $Z_i$ to become independent, as the Kullback-Leibler-Divergence encourages the posterior $q(Z|x^{(i)})$ to get closer to this distribution. While this approach successfully encourages the latent space to be disentangled the reconstruction of images gets worse. Up to the authors of [Kim and Mnih, 2019] this phenomenon appears because the Kullback-Leibler-Divergences contains the mutual information between $X$ and $Z$, as it can be decomposed as follows:

$$\frac{1}{N}\sum_{i=1}^{N} KL(q_\phi(Z|x^{(i)})||p(Z)) = I(X, Z) + KL(q_\phi(Z)||p(Z))$$

where $I(X, Z)$ denotes the mutual information between the random variables $X$ and $Z$, see Appendix 7.1, and:

$$q_\phi(Z) = \frac{1}{N}\sum_{i=1}^{N} q_\phi(Z|x^{(i)})$$

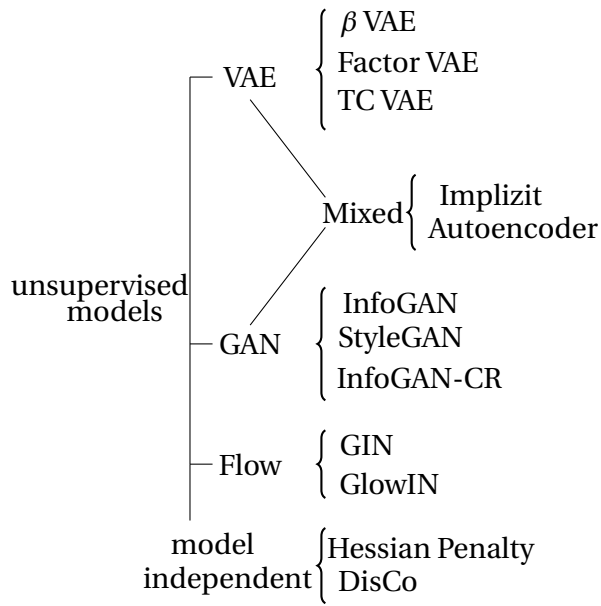

*Figure 5: Overview of the reviewed unsupervised approaches.*

Thus, weighting the KL term more decreases the mutual information between $X$ and $Z$, and the latent space representation $z$ cannot store as much information about $x$ as before. They assume the second part to be responsible for the success of the $\beta$-VAE, with respect to the degree of disentanglement, and that the decreased mutual information is the reason for the reduction of the reconstruction quality. Therefore, they introduce a model avoiding the decrease of the mutual information, the FactorVAE.

TABLE 1 Timeline: Generative Models and Metrics. The timeline considers the first online version we found, not correlated with the version which we cited.

| Model | Year | Metric |
|---|---|---|
| InfoGAN | 2016 | |
| $\beta$ VAE | | Z-diff |
| FactorVAE | 2017 | Z-min Var |
| | | SAP |
| TCVAE | 2018 | MIG |
| | | DCI |
| Implicit Autoencoder | | PPL |
| StyleGAN | | |
| InfoGAN-CR | 2019 | UDR |
| GIN | 2020 | MMT |
| Hessian Penalty | | CIFC |
| DisCo | 2021 | |
| GlowIn | | |

**FactorVAE**

The FactorVAE [Kim and Mnih, 2019] is also a Variational Autoencoder with a modified loss function. As mentioned before the authors want to avoid a decreasing reconstruction quality while encouraging a disentangled latent space representation, see Section 3.1.1. They want the posterior distributions to become independent without penalizing the mutual information between the latent space representation $z$ and the data input $x$. Therefore, they use the normal ELBO, see Equation (1), while adding a regularization term and, thus, obtain the following loss function:

$$L(\theta,\phi) = \frac{1}{N}\sum_{i=1}^{N}\left[\mathbb{E}_{q_\phi(Z|x^{(i)})}(-\log p_\theta(x^{(i)}))\right.$$

$$+ KL(q_\phi(Z|x^{(i)})||p(Z))]$$

$$+ \gamma KL(\frac{1}{N}\sum_{i=1}^{N}q_\phi(Z|x^{(i)})||\bar{q}(Z))$$

where $x^{(i)}, i = 1,..,N$ denotes a data point, $\gamma$ a constant and $N$ the number of samples and:

$$\bar{q}(Z) := \prod_{j=1}^{d} q_\phi(Z_j)$$

with the latent space dimension $d$ and $q_\phi(Z_j) = \frac{1}{N}\sum_{i=1}^{N} q_\phi(Z_j|x^{(i)})$. The term $KL(\frac{1}{N}\sum_{i=1}^{N} q_\phi(Z|x^{(i)})||\bar{q}(Z))$ is known as *Total Correlation*.
As it is difficult to sample from complex distributions $\bar{q}(x)$ or $\frac{1}{N}\sum_{i=1}^{N} q_\phi(Z|x^{(i)})$ the authors try a different approximation approach, using the *density ratio trick*. To compute the Total Correlation they introduce a discriminator $D$ returning the probability that the input $z$ was an sample from $\frac{1}{N}\sum_{i=1}^{N} q_\phi(Z|x^{(i)})$ or $\bar{q}(Z)$. They also introduce a metric, which we denote as Factor metric, see Section 4.

**$\beta$-TCVAE**

In [Chen et al., 2018b], they follow the same idea as Kim and Mnih [2019] assuming that the *Total Correlation* is the source of a disentangled representation. Hence, they introduce a uniformly distributed random index variable $n \in \{1,2,...,N\}$ with respect to the data points $x^{(n)}$ of the entire data set. Then posterior $q(Z|x^{(n)}) = q(Z|n)$ stays the same. The joint distribution of $Z$ and $n$ is given as:

$$q(Z,n) = q(Z|n)p(n) = \frac{1}{N}q(Z|n)$$

Then they decompose the expected value of the Kullback-Leibler-Divergence with respect to the dis-tribution $p(n)$ into three terms:

$$\mathbb{E}_{n\sim p(n)}[KL(q(Z|n)||p(Z))] = KL(q(Z,n)||q(Z)p(n))$$
$$+ KL(q(Z)||\prod_{j=1}^{d} q(Z_j))$$
$$+ \sum_{j=1}^{d} KL(q(Z_j)||p(Z_j))$$

With this decomposition, they claim to achieve an equivalent loss function as in [Kim and Mnih, 2019], while using the distribution of $n$ allows them to train the model with their *mini-batch weight sampling* technique, while in [Kim and Mnih, 2019] they need to train a discriminator jointly with the VAE to compute the *Total Correlation*.
The authors performed various experiments with a different weighting of all three terms of the Kullback-Leibler-Divergence decomposition. They obtained the best results by only weighting the second term more, thus, the *Total Correlation*. They also introduce a metric to measure the degree of disentanglement, which they denote as Mutual Information Gap, see Section 4.

In general, an advantage of the VAE approaches is the training stability, but their generated samples are often blurry. Thus, the image quality is often low. Other VAE-based disentanglement approaches are presented in [Mathieu et al., 2019], [Kim et al., 2019] and [Kumar et al., 2018].

### 3.1.2 GAN-based Approaches

It has often been proposed that VAE approaches attempt better disentangling results than GAN-based methods, thus, there has been more attention to VAEs at the beginning. But up to Lin et al. [2020] this was only caused by unfavorable training parameters and meanwhile, GAN approaches yield as good results as VAEs.

**InfoGAN**

The *InfoGAN* [Chen et al., 2016] model uses a normal GAN structure, but divides the latent code representation into two parts, $z$ and $c$. The part $z$ should contain all useful information, while $c$ only contains noise. This is encouraged by maximizing the mutual information between the output of the generator $G(z,c)$ and the latent code $z$. Therefore, to the usual minimax game of a GAN, a regularizer term is added:

$$V(D,G) - \lambda I(G(c,z),c)$$

where $I$ denotes the mutual information and $V(D,G)$ the GAN loss function, see Section 1.1.

**StyleGAN**

The authors of [Karras et al., 2019] introduce a GAN architecture with a modified generator network. Instead of the latent space representation, which is usually the input for a generator, the network gets a constant as input. The latent space representation is used differently, by a mapping network that transforms the latent space variable $z$ onto a variable $w$ in an auxiliary space $W$. Linear transformations of $w$ are integrated into the convolution layers of the generator. These layers also get noise variables as input. This separation should enable the network to create stochastic variations in the output from the noise variable and different styles from the transformations of $w$. The authors claim that the transformation of the latent space onto an auxiliary space is useful because the latent space input has to follow the distribution of the training data. If several attributes are missing in the data or low represented they are not represented in the latent space. To avoid sampling these 'illicit' samples the latent space is kind of curved and hence is unavoidably entangled. The transformation onto $W$ should undo this curvature and offer a more linear structure. Further research on this architecture can be found in [Karras et al., 2020] and [Lang et al., 2021].

**InfoGAN-CR**

Like in InfoGAN, for the InfoGAN-CR approach, introduced by Lin et al. [2020], the latent space representation is split into noise and informative latent factors. Like in the normal GAN approach, the architecture consists of a generator and a discriminator. But it also contains an extra discriminator $H$, called *contrastive regularizer*. This discriminator gets two input images $x'$ and $x''$. These inputs are generated from latent space vectors which share the same fixed value of dimension $i$, thus, $z'_i = z''_i$. Then the discriminator is trained to identify $i$, the index of the fixed factor. This works as follows:

They sample an index $i$ from the set of all possible dimensions $I = \{1,..,k\}$ where $k$ is the dimension of the latent space $z$, the part without noise. Then two latent space representations $z^1$ and $z^2$ are created and their *contrastive gap* is computed as:

$$\min_{j \in I \setminus \{i\}} |z^1_j - z^2_j|.$$

Next, they create two images $x^1$ and $x^2$ from $z^1$ and $z^2$ with the image generator from the standard GAN approach. These two images are fed into the discriminator $H$ which tries to identify which code was shared.

### 3.1.3 Combination of VAE and GAN

**Implizit Autoencoder**

The implicit autoencoder [Makhzani, 2019] consists of the architecture of a Variational Autoencoder but uses an additional discriminator network, see Figure 6. The author argues that normal Variational Autoencoder can only learn factorized posteriors and conditional likelihoods. Avoiding this he introduces a Variational Autoencoder that learns implicit distributions. They denote a distribution as implicit if not parameters like expected value or variance of a density function are computed but a function which generation samples of this distribution, as it is done with a generator of a GAN model. Here the encoder $enc$ gets the data input $x$ and a noise vector $\epsilon$, thus, $\hat{z} = enc(x,\epsilon)$. Then the decoder gets $z$ and a noise vector $n$ as input, thus, $x' = d(z, n)$. This separation into $z$ and noise enables the network to only save the content information in the latent space variable $z$ and generate stochastical variation (style) with the noise vector, which allows a weak form of disentanglement of this information. The discriminator $D$ learns to distinguish between positive samples $(x, \hat{z})$ and negative samples $(x', \hat{z})$.

### 3.1.4 Flow-based Approaches

**General Incompressible-Flow Networks**

Sorrenson et al. [2020] make both theoretical and practical contributions: firstly, they derive the exact conditions under which disentanglement is possible and identifiable. Hereby, they formally equate the task of disentanglement to nonlinear Independent Component Analysis (nonlinear ICA) Hyvärinen and Oja [2000]. The authors find that the task only has an identifiable solution as long as additional information is given along with the data, such as class labels, which has to fulfill certain theoretical assumptions. From their theory, they develop a new type of normalizing flow called General Incompressible-Flow Network (GIN), which is to perform identifiable disentanglement in the sense of nonlinear ICA.

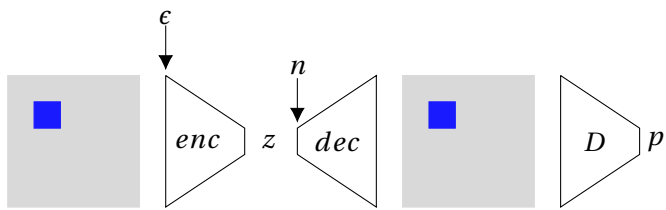

Figure 6: Normal VAE architecture, encoder $enc$ and decoder $dec$, extended by a discriminator $D$. Noise vectors $n$ and $\epsilon$ are added. The probability that the input was an positive example is denoted with $p$.

**GLOWin**

Sankar et al. [2021] take a straight-forward approach to disentanglement with normalizing flows: Essentially, they train a standard GLOW normalizing flow (see Section 1.3), and simply apply the disentanglement *factor loss* introduced by Esser et al. [2020] to the latent space. Due to the high number of dimensions and the specifics of the network architecture (namely skip connections), the factor loss is only applied to a subset of latent dimensions deemed most semantically relevant. The authors evaluate their method on the BraTS 2019 dataset of brain MRI scans [Menze and et al., 2014], demonstrating meaningful results on real-world medical image data.

In general, VAE approaches mostly try to achieve disentanglement by encouraging the distribution $q(z|x)$ to be factorized, while most GAN approaches try to achieve disentanglement by maximizing the mutual information between the latent code and the generator output. Flow-based Models often try to use the theory of Independent Component Analysis (ICA) [Jutten and Herault, 1991; Hyvärinen and Oja, 2000] to decompose the input into its independent components.

### 3.1.5 Model-Independent Approaches

It is also possible to disentangle latent space representations of pretrained models, independent of their architecture, like in [Esser et al., 2020]. Two other possible methods of model-independent techniques are Hessian Penalty [Peebles et al., 2020] and DisCo [Ren et al., 2021].

**Hessian Penalty**

The authors of [Peebles et al., 2020] describe their understanding of disentanglement as a connection between the latent space representation and the generated output. They assume a representation to be disentangled if the change of the output -while varying one latent dimension- is independent of the remaining fixed factors, see Figure 7. To achieve this, they present an approach that could be used for any generative model, by adding a regularization term to an arbitrary loss function. Therefore, remembering the generator or decoder of a model as a function that maps the latent space variables onto the data space:

$$G : \mathcal{Z} \to X$$

with $G(z) = (G_1(z), ..., G_N(z))$, $N$ the dimension of a flattened data point $x$ and $G_i : \mathcal{Z} \to \mathbb{R}, i = 1,..,N$. The first derivative of one of these functions, $\frac{\partial G_i}{\partial z_j}$, describes how much changing the latent variable $z_j$ changes the output. Now taking the second derivative of $G_i$, $\frac{\partial G_i}{\partial z_j \partial z_k}$, intuitively describes how much the change of the output is influenced by changing $z_k$. Thus, from the disentanglement point of view, all second derivatives should be zero for $k \neq j$. Hence, the authors argue that these functions, which are represented by the decoder or generator of a model, should have a diagonal Hessian matrix $H$ because the entries of this matrix are described by the derivatives:

$$H_{jk} = \frac{\partial G_i}{\delta z_j \partial z_k}$$

Thus, they suggest adding a regularizer term to an arbitrary loss function, which encourages the generator (or decoder) to consists of functions with a diagonal Hessian matrix. The term is defined as:

$$L_H(G) = \max_i L_H(G_i)$$

where $L_H(G) = \sum_{i=1}^{d} \sum_{j \neq i}^{d} H_{ij}^2$. The authors introduce an simple way to approximate the Hessian matrices as they prove the relation:

$$L_H(G) = Var_v(v^T H v)$$

where $Var_v$ denotes the variance about the Rademacher vectors $v$, which entries are $-1$ or $+1$ with equal probability. The variance can be approximated with empirical estimation, which still requires the computation of the Hessian matrix $H$ of the generator $G$. This can be done with a finite difference approximation:

$$v^T H v \approx \frac{1}{\epsilon^2}(G(z + \epsilon v) - 2G(z) + G(z - \epsilon v))$$

where $\epsilon > 0$ is a hyperparameter influencing the accuracy of the estimation of the second derivative and $z$ is the latent space input to the generator $G$.

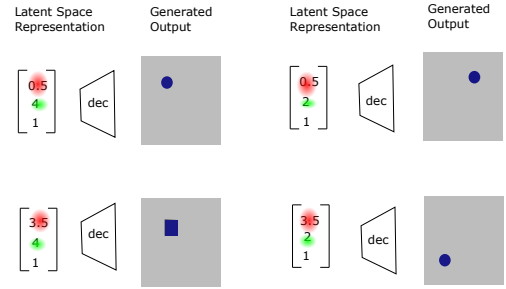

Figure 7: **Left side:** *Changing the first entry (marked red) and keeping the others fixed, changes the shape of the object in the output.* **Right side:** *For a different fixed value at the second entry (marked green), the change of the first entry does not change the shape of the object anymore but its position. Therefore the effect of the change of the first entry is influenced by the second entry.*

**Disentanglement via Contrast**

The authors of [Ren et al., 2021] assume that for any generative model, disentangled directions already exist in its latent space. That means they assume that directions $r$ contain the semantic meaningful change information. Thus, moving the latent space representation into this direction results in an interpretable change in the image. For instance, a three-dimensional latent vector would not be moved along an axis by adding a unit vector $e^i = [e_1^i, e_2^i, .., e_d^i]^T$, $e_k^i = 1$ if $i = k$, otherwise $e_k^i = 0, i, k \in \{1, 2, .., d\}$, like:

$$\begin{bmatrix} z_1 \\ z_2 \\ \vdots \\ z_d \end{bmatrix} + \epsilon \begin{bmatrix} e_1^i \\ e_2^i \\ \vdots \\ e_d^i \end{bmatrix} \text{ but by } \begin{bmatrix} z_1 \\ z_2 \\ \vdots \\ z_d \end{bmatrix} + \epsilon \begin{bmatrix} r_1 \\ r_2 \\ \vdots \\ r_d \end{bmatrix}$$

where the vector $r = [r_1, .., r_d]^T$ is the direction and the scalar $\epsilon$ describes the length of the shift, see Figure 8. To find these directions they introduce their *Disentanglement via Contrast (DisCo)* approach, where they use *Contrastive Learning* on pretrained models. They introduce a *Navigator A* [2] searching the latent space for semantically meaningful directions. They sample one direction $r$ and a latent vector $z$ from the pretrained model. Then they feed the latent vector $z$ and a shift of this vector $z = z + A(r, \epsilon)$ in direction $r$ into the pretrained generator. Thus, they obtain two images, which, if $r$ is a disentangled direction only differ in a semantic meaningful aspect. These two images are then fed into two weight shared encoders, which return lower dimensional representations. The difference between these representations is one data point in the *Variation Space* and should be a disentangled representation. To train this model they sample three latent vectors $z$, $z^+$, and $z^-$. The vectors $z$ and $z^+$ need to obtain a second sample within the same direction, while $z^-$ needs a sample from a different direction, see Figure 8. Therefore, they receive six different images:

$$G(z), G(z'), G(z^+), G(z'^+), G(z^-), G(z'^-)$$

where $G$ is the generator or decoder from the pretrained model, which stays fixed. For a disentangled direction, the change in the image between $G(z)$ and $G(z')$ should contain the same semantic change as between $G(z^+)$ and $G(z'^+)$. This should encourage positive pairs, which are images from different perspectives or augmented images, to be close together in the latent space representation and negative pairs, which are completely distinct images, to be apart.

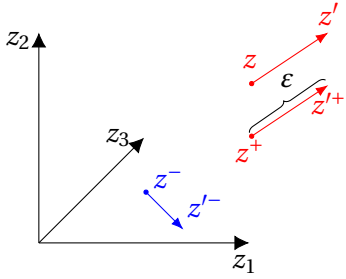

Figure 8: **Red:** Two samples with the same direction, **Blue:** Negative sample, which means samples with a different direction. If the semantic change in the generated images $G(z)$ and $G(z')$ is the same as in $G(z^+)$ and $G(z'^+)$, then this direction is a disentangled one.

---
[2] Matrix or non-linear operator of three fully-connected layers.

## 3.2 Identifiable Problem

In [Khemakhem et al., 2020], they claim that, in general, the so-called reparametrization trick of the generative models, thus, using a latent space distribution, comes along with the drawback that these models are unidentifiable. In the context of generative models, this means that different distributions $p(X|z)$ can be learned, while still fitting the data distribution as described in the following. In general, a deep latent variable model learns a joint distribution:

$$p_\theta(x, z) = p_\theta(x|z)p_\theta(z)$$

such that it fits the data distribution:

$$p_\theta(x) = \int_{\mathcal{Z}} p_\theta(x, z) dz$$

The problem with unidentifiable models is that the distribution of $p_\theta(x, z)$ could attempt different forms, while still fitting the distribution $p_\theta(x)$, see Figure 9. Thus, the model distribution does not need to approximate a ground truth distribution in the latent space but could take any form. Hence, to be able to identify the latent space distribution, they want a model to satisfy:

$$\forall (\theta, \theta') : p_\theta(x) = p_{\theta'}(x) \Rightarrow \theta = \theta'$$

where $\theta$ is a vector of parameters. They want this to be true except for simple transformations, like permutations or signed scaling. They combine the theory of non-linear Independent Component Analysis [Jutten and Herault, 1991; Hyvärinen and Oja, 2000] with Variational Autoencoders to create an *Identifiable Variational Autoencoder*, denoted as *iVAE*, which requires a conditional factorization of the latent prior distribution and thus, a kind of supervised model. Locatello et al. [2018] showed that for a general unsupervised model with a factorized prior $p(z)$, it is not possible to identify which representation is learnt. That means, assuming a model that learns a disentangled representation of the data, their exist an equivalent model which learns an entangled representation, while both models still fit the data distribution $p(x)$. But they also point out that this theoretical result does not mean that learning a disentangled representation is impossible in practice, as inductive biases on models and data can be made. In [Willetts and Paige, 2021], for example, they introduced a clustering technique for the latent space to avoid the need for labels. They showed empirically that this leads to a latent space distribution that is identifiable.

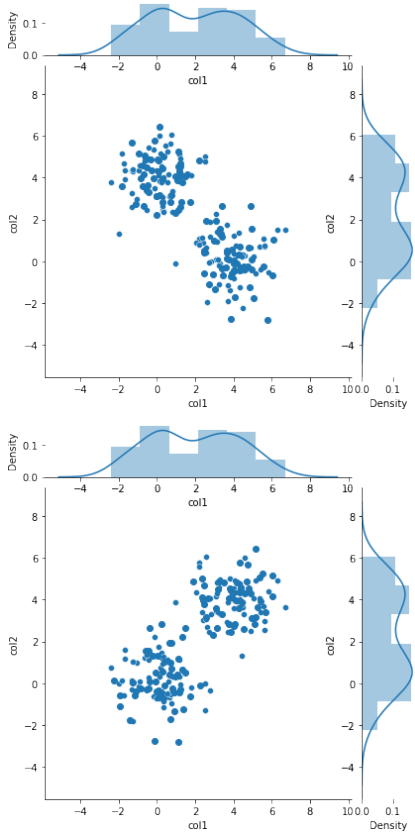

*Figure 9: Two joint probability distributions of the same marginals.*

# 4 Evaluation

As there is no explicit definition of a disentangled representation, measuring disentanglement is not consistent and still challenging. Different metrics measure different aspects of a disentangled representation, as described by Sepliarskaia et al. [2021]. They categorize the metrics into three groups, differentiating if they measure *disentanglement, completeness* or *informativeness*, with respect to the disentanglement definition of Eastwood and Williams [2018], see Section 2. Furthermore, they examine the behavior of different metrics and they reveal some failure modes, see Table 2. Up to them several metrics are not correctly scaled. This means that there exist representations completely fulfilling the measured property but achieving a low score and vice versa.

Another way to classify metrics is presented in [Zaidi et al., 2020]. Here the metrics are separated into three classes, see Table 2, up to their way to evaluate the latent space representation:

1. *Intervention-based:* compare latent space representations resulting from subsets of data generated with one fixed ground truth factor.

2. *Predictor-based:* measure the performance of a regressor or classifier, trained to predict ground truth factors from the latent space representation.

3. *Information-based:* use information theory concepts to evaluate the latent space representation.

Also, Locatello et al. [2020] analyzed the agreement of several metrics and observed systematic differences among their correlation. But this is not the only difficulty while measuring the degree of disentanglement. Some metrics are architecture-dependent. The $\beta$-VAE metric [Higgins et al., 2017], for example, needs an encoder structure to return a disentanglement score, which is not given in most GAN-based approaches. Furthermore, most metrics are only available for data sets, where ground truth factors are known. Only a few approaches try to measure the degree of disentanglement for unsupervised models, e.g. [Zhou et al., 2021]. Therefore, we separate the metrics as unsupervised and supervised, up to their need of ground truth factors. It is also worth mentioning that some metrics need to train classifiers or regressors to return a score and thus depend on the performance of these models. Hence, we also mention if the metrics need hyperparameter to be tuned or not, see Figure 10.

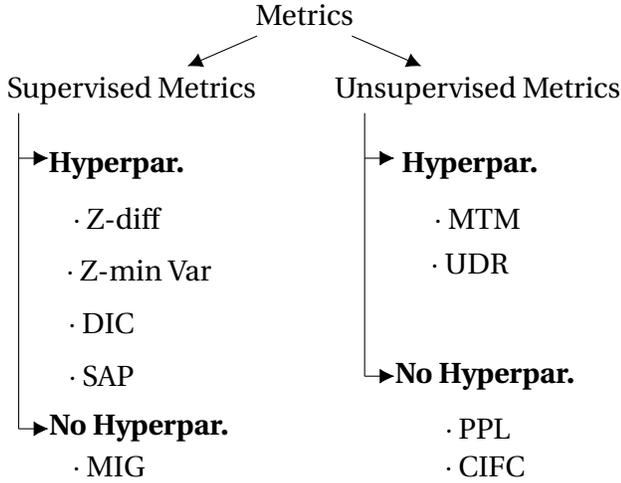

Figure 10: *We classified the disentanglement metrics in the following way: Supervised Metrics require the ground truth factors, Unsupervised Metrics do not need the ground truth factors. If the metrics are listed under Hyperparameter, regressors or classifiers have to be trained.*

**Remark:**
As image reconstruction quality is often lower for models with disentangled representations, evaluation of such an approach is often combined with an image quality metric, like in [Lin et al., 2020] with the Interception Score (IS), Frechet inception distance (FID), Peak signal-to-noise-ratio (PSNR), structural similarity index (SSIM), Contrast-to-noise ratio (CNR), Equivalent number of looks (ENL), Conditional Generation Accuracy (CGAcc) [Havaei et al., 2021].

## 4.1 Supervised Metrics

We denote metrics that need the ground truth factors to be known as supervised metrics.

### Z-diff

The Z-diff [3] metric, or $\beta$-VAE metric [Higgins et al., 2017], was one of the first approaches to evaluate a representation with respect to its disentanglement behavior. The author's idea is that for a disentangled model the latent space representation of images generated by ground truth factors with one fixed value should obtain the same value, see Figure 11.

To evaluate the degree of disentanglement, one dimension of the ground truth vector $g$ after the other is fixed and all remaining dimensions are chosen randomly. Then, multiple images $x$ and their corresponding latent space representation $z$ are generated. For a disentangled representation the dimension held fixed in the ground truth factor should be the same in the latent representation. Thus, the difference between the latent space representation and the ground truth factor $|g - z|$ is one data point for a classifier, with the fixed factor as the label. For a disentangled representation the difference $|g - z|$ should be zero at the dimension of the fixed factor. The disentanglement score is then defined as the error rate of the classifier. But this metric has a failure mode, as a classifier could classify all points correctly even if one dimension is not disentangled, thus giving a score of one to a not completely disentangled representation.

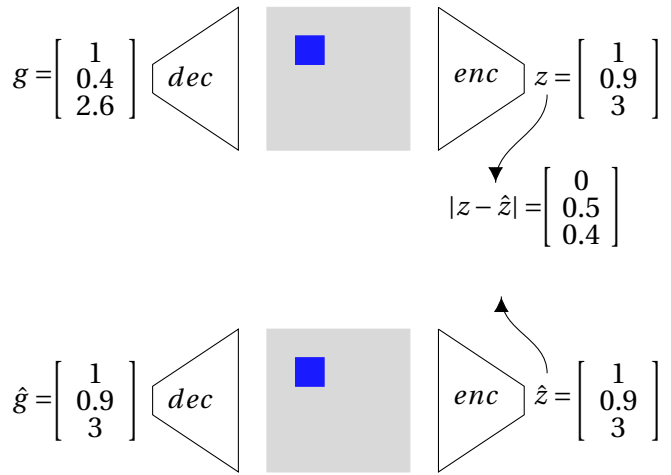

Figure 11: *Taking two ground truth vectors $g$ and $\hat{g}$ with one fixed dimension and randomly chosen remaining values, generate two images and receive their latent space representation $z$ and $\hat{z}$. The difference of them is a data point for the classifier.*

---

[3] We follow the nomenclature of [Zaidi et al., 2020] and [Do and Tran, 2021].

Table 2: Metrics: Z-diff, Z-min Var, Mutual Information Gap (MIG), Disentanglement-Compactness-Explicitness (DIC), Separated Attribute Predicatbility (SAP), Manifold Topology Metric (MTM) and Perceptual Path Length (PPL), Cross Image Feature Consistency (CIFC) error and Unsupervised Disentanglement Ranking (UDR).

| Metric | Scaled to [0, 1] | Designed to Measure | Measure Principles [Zaidi et al., 2020] | Failure Mode Detected [Sepliarskaia et al., 2021] | Supervised |
|---|---|---|---|---|---|
| Z-diff | ✗ | Disentanglement [Zaidi et al., 2020] | Intervention-based | ✓ | ✓ |
| Z-min Var | ✗ | Disentanglement [Zaidi et al., 2020] | Intervention-based | ✓ | ✓ |
| SAP | ✓ | Compactness [Zaidi et al., 2020] | Prediction-based | ✓ | ✓ |
| MIG | ✓ | Compactness [Zaidi et al., 2020] | Information-based | ✗ | ✓ |
| DCI | ✓ | Disentanglement, Compactness, Explicitness [Zaidi et al., 2020] | Prediction-based | ✓ | ✓ |
| PPL | ✗ | Latent space structure | - | - | ✗ |
| UDR | ✗ | Model Parameter Setting | - | - | ✗ |
| MTM | ✗ | Latent space structure | - | - | ✗ |
| CIFC | ✗ | Content/style separation | - | - | ✗ |

**Z-min Variance (Z-min Var)**

The Z-min Variance [4] metric, or FactorVAE metric, introduced in [Kim and Mnih, 2019] corresponds to the *Z-diff* metric but tries to avoid the failure mode. They also keep one dimension of the ground truth vector *g* fixed, generate batches of samples and their latent space representation. Then, after taking the difference $|g - z|$, they choose the dimension with the lowest variance and create a data point for a classifier with the dimension and the ground truth factor as label.

**Separated Attribute Predictability (SAP)**

Computation of the SAP score [Kumar et al., 2018] consists of two steps:

1. Construct a score matrix *S*:

   For every combination of a ground truth factors $g_i, i = 1,..,K$ with a dimension of the latent space representation $z_j, j = 1,..,d$ a score $s_{ij}$ is computed as the $R^2$ score of a regressor [5], which is a a statistical measure of the quality of the regressors prediction. The scores $s_{ij}$, which are ranged between zero and one are the entries of the matrix *S*.

2. Compute gaps:

   Then, for every column, they compute the difference between the highest and second-highest value. A large gap refers to a disentangled connection between the corresponding ground truth factor and latent space dimension. The mean of these differences is then the total SAP score.

---

[4]We follow the nomenclature of [Zaidi et al., 2020].

[5]Or a decision tree for categorical factors [Zaidi et al., 2020].

**Mutual Information Gap (MIG)**

The *Mutual Information Gap* is a metric introduced in [Chen et al., 2018b]. To measure the degree of disentanglement the mutual information between every dimension of the latent code factor $z_i$ and the ground truth factor $g_j$ is computed:

$$I_{ij} = I(z_i, g_j) \quad i = 1,...,D \ \ j = 1,...,K$$

where $I$ denotes the mutual information, see Appendix 7.1. This measures how much information about $g_j$ is stored in $z_i$. For a disentangled representation only $g_i$ and $z_i$ should have high mutual information, all others should be zero. To check this, the highest and the second-highest mutual information is taken into account. The difference between them is computed and normalized by the sum of the mutual information between all factors and latent representations. Therefore, this metric gives a score between zero and one while one is the best score.

**Disentanglement, Completeness, Informativeness (DCI)**

The *DCI* [6] metric, introduced by Eastwood and Williams [2018], consists of three separate scores, measuring *disentanglement, completeness* and *informativeness*, see Section 2. To apply this metric, the ground truth factors need to be known. Then, the following steps can be executed:

- Train the model $M$ which should be evaluated on a artificial data set with known ground truth factors $g$, e.g. dSprites.

- Receive the latent code $z$ for each sample $x$ in the data set, that means first create $x$ by the image generation part of the synthetic data set $G(g) = x$, and then create $z$ by $M(x) = z$.

- Train $K$ regressors $f_j, j = 1,..,K$ to predict ground truth factor $g_j$ given the latent code $z$, thus

$$f_j : \mathbb{R}^d \to \mathbb{R} \ \ \text{with} \ \ f_j(z) = g_j, j = 1,..,K$$

where $K$ is the dimension of the ground truth vector and $d$ the dimension of the latent space vector.

---
[6]This abbreviation was introduced in [Locatello et al., 2018].

Therefore, using LASSO or Random Forest as linear regressors, as suggested by the authors, for every ground truth factor $g_j$, an approximation is given as:

$$g_j \approx b - z^T w, \ \ j = 1,..,K$$

For simplicity we assume the bias term b to be zero. Then the absolute value of the weights $w$ of the approximation can be used as entries of the following *importance matrix*:

$$\begin{bmatrix} g_1 \\ \vdots \\ g_K \end{bmatrix} \approx \begin{bmatrix} r_{11} & \ldots & r_{D1} \\ \vdots & \vdots & \vdots \\ r_{1K} & \ldots & r_{DK} \end{bmatrix} \begin{bmatrix} z_1 \\ \vdots \\ z_d \end{bmatrix} \quad (6)$$

Scaling the weights to a range between zero and one by:

$$P_{ij} = \frac{r_{ij}}{\sum_{k=1}^{K} r_{ik}}.$$

This is interpreted as a measure of importance of $z_i$ to predict $g_j$. Then, the disentanglement score is defined as:

$$D_i = 1 - H_K(P_{i.}) \quad \text{where} \quad H_K(P_{i.}) = -\sum_{k=1}^{K} P_{ik} \log_K P_{ik}.$$

For example, if all entries of the first column in (6) would have the same value then $z_1$ would be equally important to predict $g_j, j = 1,..,K$. Hence, $P_{1j}$ would be $\frac{1}{K}$ for all $j = 1,..,K$. Then, as $\log_K(\frac{1}{K}) = -1$, we get $H_K(P_{1.}) = 1$. Thus, the disentanglement score $D_1$ for the first latent space code $z_1$ is equal to zero and, therefore, not disentangled.

The individual scores for all dimensions of the latent code are concatenated to one overall disentanglement score, computed as:

$$D = \sum_{i=1}^{d} \rho_i D_i \quad \text{where} \quad \rho_i = \frac{\sum_{j=1}^{d} r_{ij}}{\sum_{i=1}^{d} \sum_{j=1}^{K} r_{ij}}.$$

The same is done for the completeness score, but not with the columns but with the rows:

$$\tilde{P}_{ij} = \frac{r_{ij}}{\sum_{k=0}^{D-1} r_{kj}} \quad \text{and} \quad C_i = 1 - H_K(\tilde{P}_{.j})$$

where

$$H_K(\tilde{P}_{j.}) = -\sum_{k=0}^{K-1} \tilde{P}_{jk} \log_K \tilde{P}_{jk}$$

The last part of the metric is the *informativeness score*. This is defined as the average of the prediction error of the regressor. The value of all three scores is scaled, thus, gives values between 0 and 1, where 1 is the best and 0 the lowest score.

## 4.2 Synthetic Data Sets

As supervised metrics need the ground truth factors to be known, to evaluate the disentanglement performance of a model, synthetic data sets are necessary. Two populate data sets are:

**dSprites and 3D shapes**

The *dSprites* data set [Matthey et al., 2017] is an often-used synthetic data set to evaluate disentanglement performance of generative models [Lin et al., 2020; Kim and Mnih, 2019]. This data set consists of 2D images of objects generated from six ground truth factors: Color, shape, scale, rotation, position (with x and y coordinate). All possible combinations of these factors are considered once, resulting in 737280 images. The 3D shapes data set [Burgess and Kim, 2018] contains images of 3D shapes generated from six independent ground truth factors: Floor color, wall color, object color, scale, shape, orientation. Like in the dSprites data set, all possible combinations of these ground truth factors are considered once, resulting in 480000 images in total.

## 4.3 Unsupervised Metrics:

We denote metrics that do not need the knowledge of the ground truth as unsupervised.

**Perceptual Path Length (PPL)**

The metric introduced in [Karras et al., 2019] is called *Perceptual Path Length* and is based on the idea that images that are only slightly different should have a latent space representation close to each other, thus, the path between them, lying in the latent space, should be short. This metric was introduced with the StyleGAN, to measure the disentanglement performance of models without an encoder structure. The basic structure is the following: Take two samples of the latent space $z_1$ and $z_2$ and choose two points by interpolation between them, thus, choosing these points from the path through the latent space between them. Generate two images from these two points and compute a distance between these two images with a metric $d$, measuring the perceptual difference between the images. A short path is better, thus, a low score is desirable:

$$L = \mathbb{E}(d(G(serp(z_1,z_2,t)), G(serp(z_1,z_2), t+\epsilon)))$$

where $z_1$ and $z_2$ are sampled randomly from $p(z)$ and $t$ from a uniform distribution $U(0,1)$, $G$ denotes the model generator and $serp$ the spherical interpolation operator:

$$serp((z_1,z_2),t) = \frac{\sin((1-t)\theta)}{\sin(\theta)}z_1 + \frac{\sin(t\theta)}{\sin(\theta)}z_2$$

where $\theta$ is the angle between $z_1$ and $z_2$. The spherical interpolation is used for normalized representations as the resulting interpolated points are again normalized. For not normalized input linear interpolation can be used.

**Unsupervised Disentanglement Ranking (UDR)**

Duan et al. [2019] emphasize the challenges of supervised evaluation of disentangled representations. To avoid the usage of fixed ground truth factors and to offer a way to select a good hyperparameter setting for a model [7], they introduce their *Unsupervised Disentanglement Ranking (UDR)*. This system is based on the assumption that disentangled representations from a model are similar to each other up to a certain degree, while entangled ones are completely different. They claim disentangled representations of a model to be alike, despite:

- *Permutations:* Representations are similar, expect for the order of the information encoding dimensions.

- *Sign:* Only the absolute value of the representation is important.

---
[7]They concentrated on VAEs.

- *Subsetting:* Does not matter how many dimensions encode an factor.

To evaluate if two representations are similar, apart from these three aspects, Higgins et al. [2021] compute a score for two representations as follows:

1. Compute the informativeness matrix $R$ like for the DCI score [Eastwood and Williams, 2018], using a LASSO regressor [8]. Considering such a matrix makes the score insusceptible for permutations.

2. Taking the absolute value of all these entries makes the score robust against different signs.

3. To address the subset aspect they divide the score by the number of informative latent dimensions, whereby they describe a latent dimension as informative, if the posterior distribution $q(z_i|x)$ diverges from the prior $p(z_i)$ and otherwise as uninformative:

$$I_{KL}(a) = \begin{cases} 1 & KL(q_\phi(z_a|x)||p(z_a)) > 0.01 \\ 0 & \text{otherwise} \end{cases}$$

The number of informative latents for model $a$ and accordingly for model $b$ are defined as:

$$d_a = \sum_{i=1}^{d} I_{KL}(z_i) \text{ and } d_b = \sum_{j=1}^{d} I_{KL}(z_j)$$

They want to assign a score not only to the model itself but also to its hyperparameter settings. Therefore, they choose $H$ hyperparameter setting for a model. Then, all models are trained with $S$ different initial states, resulting in $H \times S$ trained models. Next, for each of the $S$ models within a hyperparameter setting, $P \leq S$ other models within this setting are chosen and pairwise compared. This pairwise comparison is done by computing the $UDR_{a,b}$ score for two models $a$ and $b$ with the same hyperparameter setting but different initialization is computed as:

$$UDR_{a,b} = \frac{1}{d_a + d_b} \Big[ \sum_{j=1}^{d} \frac{r_i^2 I_{KL}(z_j)}{\sum_{i=1}^{d} R(i,j)} + \sum_{i=1}^{d} \frac{r_j^2 I_{KL}(z_i)}{\sum_{j=1}^{d} R(i,j)} \Big]

---
[8]The authors of [Higgins et al., 2021] also introduce a non-parametric informativeness matrix with the Spearman Correlation.

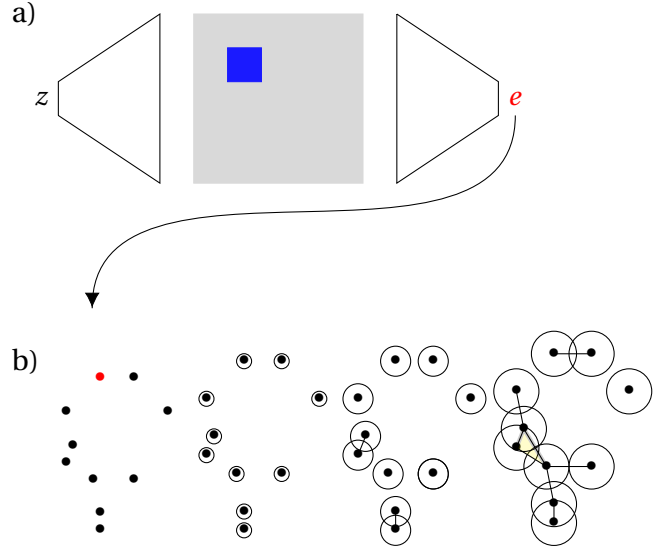

Figure 12: a) For every latent dimension $z_i$, images are generated with the model to be evaluated, then, these generated images are fed into a network, e.g. a VG16 [Simonyan and Zisserman, 2014], to obtain an image embedding. b) These embeddings are data points, which are transformed into a graph. Every data point represents a node and is surrounded by a ball whose radius increases. If two balls overlap, the corresponding points are connected through an edge. Then, a graph with holes arises. But these holes will also disappear with increasing radii. Therefore, the living time of these holes is calculated. The Relative Living Time (RLT) is a discrete distribution describing how many holes exist at what time. These distributions are used for clustering. Therefore, the Wasserstein distance between distributions is used.

where $R(i,j)$ is the entry of the informativeness matrix $R$ at row $i$ and column $j$, $r_i = \max_j R(i,j)$ and $r_j = \max_i R(i,j)$.

**Manifold Topology Metric (MTM)**

A new unsupervised metric is introduced by Zhou et al. [2021] and we denote it here with MTM. They built their metric upon the manifold hypotheses. According to this hypothesis, the data lie on a manifold which the generative model tries to approximate. They set up a disentanglement score by comparing the topology of the submanifolds conditioned on a latent dimension $z_i$. They measure the similarity of the submanifolds that are created by $p(x|z_i = v)$ where $z_i$ is the $i-th$ dimension of the latent space representation and $v$ is a value that is changed to create a set



of data points. They assume if the generative model is disentangled, then these submanifolds have different topologies. To identify the topology of the submanifolds they use a graph structure, see Figure 12.

**Cross Image Feature Consistency (CIFC) error**

Havaei et al. [2021] introduce a metric to measure how a latent representation that consists of two parts separately contains the information of two factors. The separation of these factors, which they denote as content and style, has to be consistent across all instances, and the representations should not share information about the input. To measure these aspects they introduce the Cross Image Feature Disentanglement (CIFC) error. Therefore, they take to images and compute their content and style representation. Then they mix these representations, thus, the content representation from the first image is combined with the style representation of the second one and the style representation of the first one is combined with the content representation of the second one. Then two new images are generated from these mixed representations. These new images are again encoded into their content and style representation and the representations are mixed again. For perfect disentangled and information preserving representations, the images which are generated from these second mixed representations would be the original input, thus their encoding should be the same as the first one and thus the difference between these encodings is taken as the basis for the score, which is the expected value of this difference about all images in the test data set.

## 5 Medical Applications

Deep Neural Networks (DNNs) are commonly used in healthcare, e.g. for image segmentation, classification, reconstruction, or image synthesis [Ronneberger et al., 2015; Wang et al., 2020; Li et al., 2021] and several other applications [Nensa et al., 2019]. The added value of a disentangled representation for medical purposes can be multifaceted:

- *Downstreaming Tasks:* Especially in healthcare, annotated data is raw and labor-intensive. Therefore, pretraining models in an unsupervised fashion with unlabeled data and specifying on a few annotated data is a popular approach. This process could benefit from disentangled representations, as task specification could be done faster.

- *Interpretability:* For clinical usage DNNs need to be reliable and trustworthy. Understanding the decision process and outgrowing the black box characteristic is necessary. Furthermore, models for medical purposes are often trained and tested on data from the same hospital. Applying these models to data from different hospitals often comes along with poor results [Zech et al., 2018]. A disentangled representation could help to detect learned shortcuts.

- *Synthetic Data Generation:* With a controllable image generation process, task-specific synthetic data could be created and analyzed, reducing the need for annotated data.

- *Digital Signatures:* Comparable to the radiomics approach [Gillies et al., 2016], a disentangled latent space representation could be used to characterize or classify tumors, avoiding the need for handcrafted features.

These are just a few possible applications, but in fact interests in medical applications of disentanglement seem to increase, see Figure 13. Therefore, we review existing medical applications of disentanglement.

**Search Criteria:**

The reviewed papers are the results found on PubMed in the period 01/01/2017 to the 01/13/2022 with the search term *Disentanglement* restricted to title and abstract. The number of resulting papers has been 219. Screening title and abstract for including machine learning issues reduced them to 79. Concentrating on medical applications and generative models reduced the number to 27 papers. However, we add five papers from other sources, see Figure 14. Furthermore, we would like to mention that different disentanglement approaches appeared with our search string, e.g. [Robinson et al., 2021] but do not consider generative models according to our notion. We give an overview about the results in the following Table 3.



Table 3: Compact overview of all reviewed studies, showing the used data sets and if these are publicly available or not, and the modalities of the data sets. Further, the table shows the underlying network architectures of the studies and what kind of supervision, e.g. supervised, unsupervised, weakly-supervised or self-supervised, etc., was applied. Finally, the table provides the information about the disentanglement factor for all reviewed works and what kind of metrics have been used for evaluation.

| Study / Application[9] | Data[10] | Modality | Supervision[11] | Network-Architecture[12] | Disentangled Factor[13] | Metrics[14] |
|---|---|---|---|---|---|---|
| **Brain** | | | | | | |
| [Higgins et al., 2021] | Face Data Set [15] | grayscaled images | self-supervised | $\beta$-VAE | complete | UDR score Human Raters |
| [Ouyang et al., 2021] | NCANDA [16] BraTS2020 [19] Zero-Dose | MRI MRI/PET | self-supervised | Autoencoder-based | anatomy and modality | PSNR SSIM |
| [Fei et al., 2021] | BraTS2015 [17] | MRI | self-supervised | GAN-based | modality shared and specific code | NRMSE PSNR SSIM |
| [Zhao et al., 2021] | Synthetic Data Set ADNI1 [18] Alcohol Data Set | MRI | self-supervised | Autoencoder-based | brain age | - |
| [Liu et al., 2021b] | BraTS2018 [19] | MRI | self-supervised | GAN-based, Autoencoder | modality shared and specific code | IS PSNR SSIM |

---

[9] Some methods consider multiple *organs*, but we classify them only under one application.

[10] We only mention Data Sets relevant for medical context. Sometimes only subsets, modifications or combinations of the mentioned data sets are used. If no footnote is adapted, the data sets are private or provided by the authors of the listed application.

[11] If the authors do not explicitly mention, if they use a supervised method or not, we denote every method using labels in any way (survival prediction, conditioning, etc.) as supervised. Except for methods which only need modality labels.

[12] We only mention the basic structure. Most networks use combinations of several networks and complex architectures.

[13] We denote the disentanglement as complete if the method considers the whole latent representation, otherwise we mention the explict factors.

[14] We only mention disentanglement or image quality metrics.

[15] Combination of CelebA [Liu et al., 2015], Chicago Face Database [Ma et al., 2015], CVL [Peer, 1999], FERET [Phillips et al., 1998], MR2 [Strohminger and et al., 2016],PEAL[Gao and et al., 2008]

[16] [Zhao et al., 2020]

[17] [Menze and et al., 2015]

[18] Alzheimer's Disease Neuroimaging Initiative: http://adni.loni.usc.edu/

[19] [Menze and et al., 2014]



| [Hu et al., 2021] | UNC/UMN Baby Connectome Project Data Set [20] | MRI | supervised | Adversarial Autoencoder | modality shared and specific code | - |
|---|---|---|---|---|---|---|
| [Hu et al., 2020] | UNC/UMN Baby Connectome Project Data Set [20] | MRI | supervised | Autoencoder-based | age identity noise | - |
| [Xia et al., 2020] | ISLES [21] BraTS2018 [17] Cam-CAN [22] | MRI | (semi-) supervised unsupervised | CycleGAN Autoencoder | pathology info from healthy | Human Raters Own image quality metrics [23] |
| [Zhao et al., 2019] | Brain Data Set | MRI | supervised | VAE-based Regression | age factor | rMSE |
| **Lung** | | | | | | |
| [Havaei et al., 2021] | LIDC-IDRI [24] HAM10000 [25] | CT RGB images | supervised | GAN-based (Cycle - and Bidirectional- GAN) | content and style | CIFC FID IS CGAcc |
| [Xiu et al., 2021] | COVID-19 Data Set InP Framingham Study [26] SEER [27] SLEEP [28] | CT | supervised | VAE-based Regression | complete | - |
| [Toda et al., 2021] | Lung Cancer Data Set | CT | supervised | InfoGAN-based | lesion shape | FID |
| [Song et al., 2020] | Lung Cancer Data Set | CT | self-supervised | BigBiGAN | complete | Compare to Radiomics |

---

[20] [Howell and et al., 2019]
[21] [Maier et al., 2015]
[22] [Taylor et al., 2017]
[23] Measure *Identity*, *Healthiness* and *Deformation Correction* of the pseudo-healthy images.
[24] [Armato III et al., 2011]
[25] [Tschandl et al., 2018]
[26] [Mitchell et al., 2010]
[27] [Ries et al., 2007]
[28] [Quan et al., 1997]



| | | | | | | |
|---|---|---|---|---|---|---|
| [Chen and Batmanghelich, 2020] | COPD Data Set | CT | weakly-supervised | VAE-based | complete | MIG |
| **Heart** | | | | | | |
| [Gyawali et al., 2021] | SimECG SimECG-torso ECG Data Set | ECG | unsupervised | $\beta$-VAE-based | complete | $\beta$-VAE metric |
| [Gyawali et al., 2019] | ECG Data Set | ECG | unsupervised | Autoencoder-based LSTM Regression | inter-subject variation | - |
| [Van Steenkiste et al., 2019] | MIT-BIH Arrhytmia Data Set [29] | ECG | unsupervised | $\beta$-VAE | independent base beats | - |
| **Liver** | | | | | | |
| [Kleesiek et al., 2021] | Liver Data Set LiTS [30] | CT | unsupervised | Implicit Autoencoder-based | digital signatures | Human Rater |
| [Yang et al., 2019] | LiTS [31] Liver Data Set | CT MRI | supervised | GAN-based Autoencoder | anatomy and modality information | - |
| [Ben-Cohen et al., 2019] | Liver Data Set | CT | supervised | Autoencoder-based | specific and unspecific factors | - |
| **Prostata** | | | | | | |
| [Shen et al., 2021] | BraTS2018 [17] ProstateX [32] | MRI | supervised | GAN-based | skeleton and flesh | NRMSE SSIM PSNR |
| **Cell data** | | | | | | |



---

[29] [Moody and Mark, 2001]
[30] [Bilic et al., 2019]
[31] [Christ et al., 2017]
[32] [Litjens et al., 2014]

| | | | | | | |
|---|---|---|---|---|---|---|
| [Yu and Welch, 2021] | Tabula Muris [33] Sci-Plex [34] Pancreatic Endocrinogenesis [35] | RNA-seq | self-supervised | VAE/GAN combination | complete | MIG FactorMetric Spearman Correlation IS |
| [Kompa and Coker, 2020] | Cancer Atlas [36] | RNA-seq | (semi-) supervised, unsupervised | VAE/GAN combination | domain information | - |
| [Bica et al., 2020] | Zebrafish Cells [37] Human Cells [38] Human Pancreatic cells [39] | RNA-seq | unsupervised | VAE-based | complete | - |
| **Other Medical Applications** | | | | | | |
| [Moghadam et al., 2022] | Mitosis Atypia Data Set [40] Camelyon16 [41] DigestPath [42] | whole slide images | supervised | GAN-based | color and structural | - |
| [Yao et al., 2022] | HaCAT [43] | CLSM | unsupervised | GAN-based | content and style | - |
| [Lee et al., 2021] | Deep Lesion Data Set [44] Spineweb [45] Dental CT Data Set | CT | unsupervised | CycleGAN-based | metal artifact artifact free | PSNR SSIM |

---

[33] [Consortium and et al., 2018]
[34] [Srivatsan et al., 2020]
[35] [Bastidas-Ponce et al., 2019]
[36] [Ramos et al., 2020] [Weinstein et al., 2013]
[37] [Athanasiadis et al., 2017]
[38] [Velten and et al., 2017]
[39] [Muraro et al., 2016]
[40] [Roux et al., 2014]
[41] [Bejnordi et al., 2017]
[42] [Li et al., 2019]
[43] [Kromp et al., 2020]
[44] [Yan et al., 2018]
[45] http://spineweb.digitalimaginggroup.ca/



| [Huang et al., 2021] | Clinical Data Sets [46] [47] | OCT | unsupervised | GAN-based | noise and content | CNR EPI MSR ENL |
| [Kalinin et al., 2021] | Protein Data Set | AMF | supervised | VAE-based | particle rotation | - |
| [Zhang et al., 2020] | CASIA-B [48] USF [49] FVG | Videos (RGB image sequence) | supervised | Autoencoder + LSTM | appearance, canonical, and pose features | - |
| [Liao et al., 2020] | Deep Lesion Data Set [44] Spineweb [45] CBCT Data Set | CT | unsupervised | GAN-based Autoencoder | content and artifact information | PSNR SSIM |
| [Polykovskiy et al., 2018] | ZINC Database [50] | SMILE | semi-supervised | Conditional Adversarial Autoencoder | protein properties | Mutual Information (not MIG) |



---

[46] [Abbasi et al., 2018]
[47] [Justusson, 1981]
[48] [Yu et al., 2006]
[49] [Sarkar et al., 2005]
[50] Zinc[12]; University of California, San Francisco, http://zinc. docking.org/subsets/clean-leads.

**Development of publication numbers**

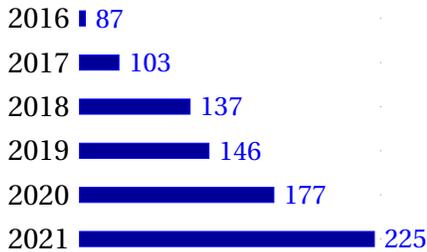

2016 — 87
2017 — 103
2018 — 137
2019 — 146
2020 — 177
2021 — 225

*Figure 13: Numbers of publications on `pubmed.gov` over the last years with respect to the search string disentangle generative model within the full text.*

We organized the reviewed papers according to the organs they considered. Some of them considered more than one, but we just separate them by one of the applications.

## Brain

### Neuron Activity

The inferiotemporal (IT) cortex of the human brain has a crucial role for the recognition of faces. In this area of the brain, faces seem to be represented by a low-dimensional neural code [Higgins et al., 2021]. To understand how such a representation is learned, Higgins et al. [2021] compare the response of neurons in the IT cortex of macaques, while presenting images of faces to them, with the latent space representation of a $\beta$–VAE [51], trained on the same data set of faces. They observed that some dimensions of the latent space representation disentangled information like hairstyle, age face shape, or smile and could explain the activity of some neurons, being sensitive to these variations. Furthermore, they showed that this disentangled representation contained less information than the encoding of other models, but offers an interpretable representation, which could be a way to comprehensive networks avoiding the black box characteristic.

### Multi-Modal Brain Analysis

Ouyang et al. [2021] introduce a network to disentangle anatomical and modality information. There-

---
[51] They also apply other generative models.

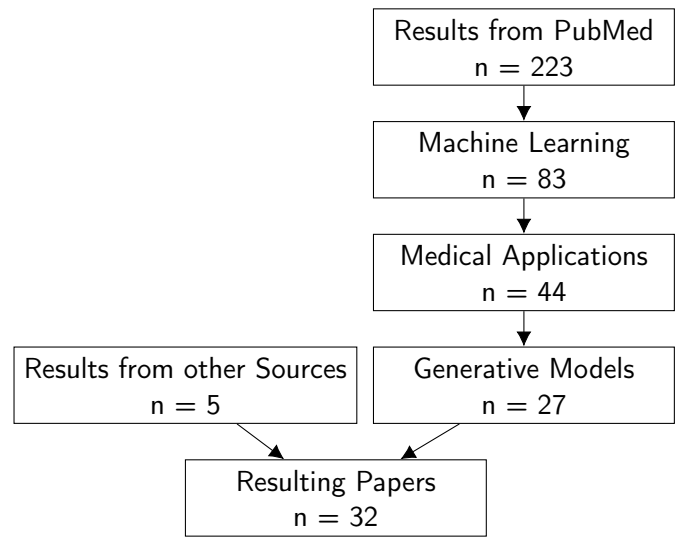

*Figure 14: Screening flowchart of the publication selection process about disentanglement approaches for medical imaging adapted from the PRISMA flow diagram by [Moher et al., 2009].*

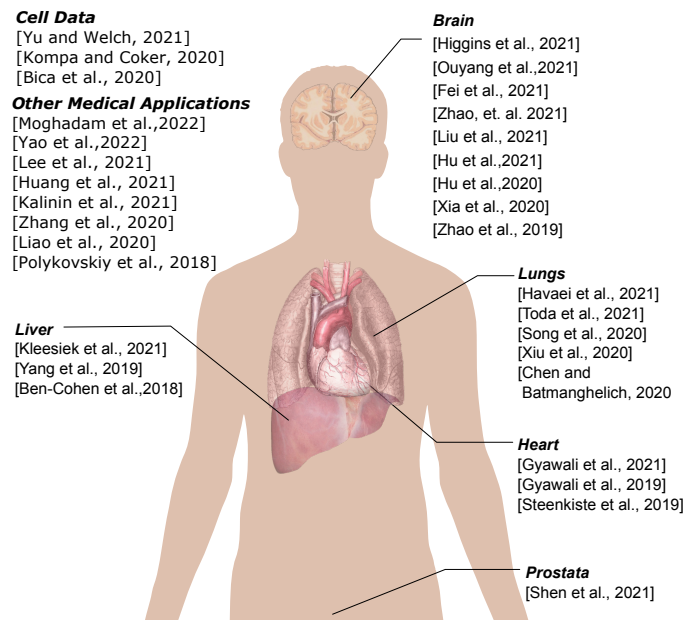

*Figure 15: The figure shows the screened medical applications that have been targeted with disentanglement (image modified from a Human body diagram).*

fore, they use and encoder to get a modality representation and an an encoder to get the anatomical representation, trained with a self-reconstruction and a cross reconstruction loss. To structure the latent space they additionally use a latent consistency loss. They mention that several works before tried ap-



proaches like this, but the disentanglement has not been checked. They give the example that both representations could contain exactly the same information and the decoder could reconstruct the input from both. To avoid this they introduce a similarity loss. They assume that the anatomical latent representation for different modalities should be similar while the modality representation should vary between the modalities. Therefore, they encourage the cosinus distance (the angle between two vectors) between the anatomical representations to be small and the one between the different modality representations to be high.

**Synthetic MRI Modality Generation**

For Magnetic Resonance (MR) images, different contrast acquisitions show different aspects, important for diagnosis. While taking all scans is time-consuming, often some image modalities are missing. The work of Fei et al. [2021] addresses this problem, by predicting missing MR image modalities of brain scans. Therefore, a GAN model with two additional networks is introduced. The input of the GAN generator is a latent representation, received of the concatenation of the information shared by all MRI modalities and the specific information of the missing modality. As the missing modality is not available during test time, a Local Adaptive Fusion (LAF) module is introduced to create a pseudo target. This network, with the disentanglement of shared and specific information, is claimed to generate higher quality images than the compared GAN models.

**Brain Age**

Zhao et al. [2021] want to analyze the effect of aging on the brain considering longitudinal MR images. Therefore, they introduce a model called Longitudinal Self-Supervised Learning (LSSL). This network consists of an Autoencoder combined with a cosine loss in the latent space, where they separate the brain age from the remaining latent representation. They can apply self-supervised disentanglement because they assume that the age of the brain is the most relevant factor influencing the brain morphology of a patient among the longitudinal scans while other factors like e.g. gender do not change over time. To disentangle the age factor, they take the difference between the latent representations of two scans of a patient at different times as a direction in the latent space. This direction represents a change in the image, which is only allowed in the direction $\tau$, which is learned while training and represents the disentangled age factor.

**Generating MRI Modalities**

Liu et al. [2021b] want to generate missing Magnetic Resonance (MR) modalities and apply these synthetic images to improve tumor segmentation tasks. They test their network on BrATs 18, with four MR image modalities. Given one modality they predict the three other modalities and connect them all as input for a standard segmentation network. To disentangle the modality information in the latent representation of the generator network, they train an encoder to be invariant to different modalities, by using different modalities as input resulting in a similar latent space representation, while the decoder gets additional information about the modality which should be generated. With the disentanglement strategy, they attempted better generated image quality than the compared baseline methods. Furthermore, the usage of these generated images for the segmentation task improved the Dice score compared to segmentation performance with synthetic images generated by other networks.

**Infantil Brain Age Prediction**

The authors of [Hu et al., 2021] modify an adversarial Autoencoder (AAE) [Makhzani et al., 2016], which they denote as disentangled multimodal adversarial Autoencoder (DMM-AAE), to predict the brain age of children of the age between birth and two years. They use multimodal input, consisting of sMR images and fMR images, and also offer an imputation strategy for missing modalities. They separate the latent space into information that all scans have in common and into scan-specific information. The latent space representations are then combined as input for a regressor to predict the age of the child. Fusion models of fMR and sMR images often end up with worse results than models using only sMRI. The authors claim that their disentanglement strategy helps to combine the information from both modalities and results in a better MAE, than AAE without this strategy.



**Infant functional connectome Fingerprinting**

The authors of [Hu et al., 2020] introduce a Triplet Autoencoder to investigate if children's brains do have functional connections that are uniquely for every individual such as a fingerprint. Therefore, they used MRI scans of the brain of children. The latent space representation is disentangled into an age, an identity, and a noise part. The age part is computed by regression, the identity by reconstruction, and the noise by adversarial learning. The network is denoted as Triplet Autoencoder because it needs three data points as input, two from the same patient at different times and one from another patient.

**Pseudo Healthy Images**

The authors of [Xia et al., 2020] want to generate *pseudo-healthy* images from pathological scans. This means they want to predict how the image would look like if the patient would be healthy. Therefore, they want to disentangle pathological and anatomical factors by training three networks together: a segmentor (U-Net [Ronneberger et al., 2015]) to detect pathological tissue, a generator to generate healthy images, and a reconstructor. For the inference process, only the generator and segmentor are needed. This is a supervised method but they also provide a semi-supervised approach. They evaluate the image quality by human raters and tested their network on three data sets: Ischemic Stroke Lesion Segmentation challenge 2015, Multimodal Brain Tumor Segmentation Challenge 2018 (BraTS) [Menze and et al., 2014] and the Cambridge Centre for Ageing and Neuroscience (Cam-CAN) data set [Taylor et al., 2017].

**Age Prediction from Structural MR Images**

The authors of [Zhao et al., 2019] want to predict the age of patients based on their structural Magnetic Resonance (MR) images. Therefore, they combine a Variational Autoencoder with a supervised regressor and disentangle the age factor in the latent space representation. The MR image is the input for both, the Variational Autoencoder and the regressor. The output of the latter is taken as a condition on the latent space of the VAE. This approach enabled a more precise prediction compared to a standard feed-forward regressor network and, through the disentangled age factor in the latent space, offers a way to visualize developmental schemes of the brain appearance by generating images with the Variational Autoencoder.

**Lung**

**CT Images and Dermatoscopic Images**

Havaei et al. [2021] introduce a GAN-based network applied to generate CT or dermatoscopic images. Thereby, they disentangle the latent space into content and style and control these factors while image generation. They denote the factors of variations conditioned on e.g. a class label as *content* and the conditioning independent factors of variations as *style* and denote their model as *Dual Regularized Adversarial Inference (DRAI)*. They apply their model on CT scans of the Lung Image Database Consortium image collection (LIDC-IDRI) [Armato III et al., 2011] and onto Human Against Machine (HAM10000) [Tschandl et al., 2018] including dermatoscopic images of seven types of skin lesions. They showed that this disentanglement approach achieved more control over the generated image and that their network achieves a better disentanglement score than extensions of an InfoGAN [Chen et al., 2016] among others.

**COVID-19 Mortality Prediction**

Medical data sets are often unbalanced, e.g. tumor size, compared to the whole body volume or mortality rates compared to survival probability. In [Xiu et al., 2021], the authors denote these cases as low-prevalence scenarios and address them with different examples like the mortality rate of COVID-19 patients, using data from the Duke University Health System (DUHS). They hypothesize that rare events are represented by extreme values of the latent factors and, thus, heavy-tailed distributions should be considered. To predict the mortality probability of COVID-19 patients, they combine a Variational Autoencoder framework, the extreme value theory, and a regressor, which helps to disentangle the influence of each latent dimension on the output likelihood. They denote their network as Variational Inference with Extremals (VIE) and claim to achieve better generalization and interpretability compared to regressors like LASSO.



**Synthetic CT Image Generation**

The authors of [Toda et al., 2021] use a modified version of an InfoGAN [Chen et al., 2016] and a WassersteinGAN [Arjovsky et al., 2017] to generate synthetic images of lung cancer lesions. They denote their approach as a semi-conditional InfoGAN because they add a third input to the generator, an additional vector including the information of the histological type of the lesion which should be generated. Then, they compare the results of a classifier pretrained on the generated data from the InfoGAN or on the generated data from the WassersteinGAN and then finetuned on the original data. In general, the classification accuracy improved for both variants, compared to a classifier without pretraining. Unfortunately, all of the latent code variables of the InfoGAN contain information about the chest wall, which limits the effect of the factor controlled image generation. The authors hypothesize that this is up to the fact that most of the training data include the chest wall and their limited latent space dimension. Nevertheless, they can control the size of the generated lesion.

**Survival of Patients with Lung Cancer**

The Authors of [Song et al., 2020] analyze CT scans from patients with non–small cell lung cancer (NSCLC), to figure out characteristics of patients who could benefit from a specific therapy. They use a BigBiGAN [Donahue and Simonyan, 2019] based network to extract semantic meaningful latent representations. They apply a regression method to extract semantic meaningful features to give a score to a patient to separate them into two groups with respect to their survival probability. With this meaningful latent representation, the results indicate better predictions of survival compared to a previous radiomics approach [Song et al., 2018], without the need for handcrafted features.

**COPD Severity**

In [Chen and Batmanghelich, 2020] the authors want to find signs for the severity of COPD disease of a patient in its Computed Tomography (CT) scans. Therefore, they introduce a weakly supervised approach of a Variational Autoencoder, where two scans are labeled with a binary label or a real-valued label describing the degree of similarity of the scans. In the latent space representation of the VAE, they want to identify factors that are related to the severity of the disease. They separate the latent representation into two parts, one to capture the information of interest and one for the remaining. Then, like in $\beta$-VAE, they regularize the Kullback-Leibler-Divergence in the ELBO, see equation (1), but only for the distribution of the latent representation part which captures the relevant information. The ground-truth factors are not available for the real world COPD data set and, thus, most metrics are inapplicable. Therefore, they apply a regressor on the latent space representation to predict several factors, relevant for COPD severity. They evaluated the prediction of the regressor with the $R^2$ score and the Cohen's kappa coefficient, achieving the best results compared to $\beta$-VAE Factor VAE and TCVAE.

**Heart**

**Anatomical Factor Disentanglement**

Electrocardiogram (ECG) signals are influenced by several factors like conduction properties of the heart, electrode positioning or heart anatomy. In the paper of Gyawali et al. [2021], they were able to disentangle five factors of variations of the anatomy of the heart. Therefore, they use a Variational Autoencoder where they generate the latent space distribution with an Indian Buffet Process [Griffiths and Ghahramani, 2011] instead of a factorized Gaussian and therefore denote their network as IBP-VAE. They encourage the disentangling of the latent space like in $\beta$-VAE by giving more weight to the Kullback-Leibler-Divergence. Furthermore, they introduce a synthetic data set of ECG signals generated from several factors that enable them to evaluate their model on that data set with the $\beta$-VAE metric. They analyze the effect of the disentangled representation on a downstream task. They train a classifier to localize the origin of ventricular activation, predicting one of 10 anatomical segments of the left ventricle (LV) of the heart. The IBP-VAE achieved better results than the compared Convolutional Neural Network (CNN).



**Ventricular Activation Origin**

Gyawali et al. [2019] want to predict the location of the origin of ventricular tachycardia (VT) from time-series ECG data. To deal with this sequential data, they introduce a network that consists of a Recurrent Neural Network, LSTM or GRU, followed by a sequential Autoencoder, whose latent space representation is separated into two parts. One part should catch the information about the location of the ventricular activation, while the other part should represent patient-individual variations. This separation is achieved by applying a contrastive loss in the latent space.

**ECG Beat Classification**

[Van Steenkiste et al., 2019] applied the $\beta$–VAE network onto ECG signals, to create a human interpretable embedding that can be used to classify the heartbeats as normal or paced. They compared the $\beta$–VAE to a normal Autoencoder. As an Autoencoder is not a generative model, traveling along the axis of the latent embedding often results in invalid outputs. For the latent space embedding of the $\beta$–VAE, they found two meaningful and human interpretable dimensions, while the remaining dimensions were not significantly relevant for the reconstruction. The meaningful dimensions encode basic shapes of a beat, which combinations can represent any beat. Therefore, they claim that classifiers, using this interpretable embedding are no longer a black box.

**Liver**

**Liver Patch Generation**

In [Kleesiek et al., 2021] they apply an implicit Autoencoder [Makhzani, 2019] to patches from Computed Tomography (CT) scans of liver lesions. In the latent code representation, which they denote as *digital signatures* of the lesions, they are able to control the spatial location and the tumor encoding information. Manipulating the latent code by replacing the tumor code part from a patch with a lesion, with the one from a patch without lesion and vice versa let them generate synthetic data. Furthermore, interpolation between the tumor code part of a patch with lesion and one without lesion and vice versa gives them the opportunity to control the appearing or disappearing of a lesion. Additionally, they train classifiers on the latent code, predicting if the input patch shows a liver lesion or not, supporting their hypothesis that the latent code can be seen as digital signatures of a lesion, offering an alternative to popular radiomics approaches [Gillies et al., 2016].

**Liver Segmentation**

The authors of [Yang et al., 2019] introduce a network denoted as Domain-Agnostic Learning with Anatomy-Consistent Embedding (DALACE) with the goal of a representation that is invariant to different modalities but contains anatomical structure information. This means they want to disentangle the anatomy and the modality information in the latent space representation. They apply their model to a domain adaption task and a domain agnostic learning task. For the domain adoption, they train the network on Computed Tomography (CT) scans with liver segmentation masks and Magnetic Resonance (MR) images without annotations, while the test procedure is only done with MR images. For the domain agnostic learning task, they train the network on CT scans with liver segmentation masks and multiphase MR images without annotations, testing only on multiphase MR images. For both tasks, the Dice score increased if the disentanglement modules have been removed. In their network they use a Domain Agnostic Module consisting of two discriminators, distinguishing between real MR or CT images and generated ones, and an Anatomy Preserving Module, consisting of a U-Net segmentation module and a discriminator. They compare their methods to several baselines, including CycleGAN [Zhu et al., 2017] and gain better results, according to Dice scores.

**Liver Lesion Classification**

Ben-Cohen et al. [2019] claim to be the first ones applying disentanglement to medical images. They create synthetic images used for data augmentation to improve the classification process of liver lesions in Computed Tomography (CT) scans. Therefore, they separate *specific* and *unspecific factors* of liver lesions. This means they separate individual characteristics of the lesions and general information like background. This is done by first training a convolutional classifier. For the second training step, this classifier is held



fixed and an adversarial classifier is trained, whose output is concatenated with the output of the first classifier. This concatenated vector is the input for a generator, reconstructing the original scan. With this separation of the generator's input, they can create new synthetic data by replacing either the specific latent representation part with one from another scan or the unspecific one. They then include their synthetic data in the training process to improve the classification prediction.

## Prostate

### Missing MRI Modalities

The authors of [Shen et al., 2021] use a GAN-based network with a disentanglement modification to separate the information which they denote as *skeleton* and *flesh*, independent of the modality. They want to generate missing modalities as well as segmentation prediction. The network consists of a content encoder getting all modalities as input, to learn the shared anatomical information, while every modality is used separately for its own style encoder, catching the modality-specific information. They compare their results to images generated from other GAN-based methods, claiming to achieve better image quality according to relevant details. They assume that this is achieved by a better content code, which is necessary to maintain the anatomical structures of medical scans. They test the network on the data sets BraTS [Menze and et al., 2014] and ProstateX [Litjens et al., 2014], which both contain MR images.

## Cell Data

### Drug Treatment Response

Tumor cells behave differently from normal cells, which is a cause of different gene expressions. To distinguish gene expressions of normal and mutated cells, the authors of Yu and Welch [2021], introduce a combination of a Variational Autoencoder (VAE) and a Generative Adversarial Network (GAN) to sample from a disentangled representation without violating data generation quality, as it usually occurs in VAEs. They apply their network to single-cell gene expression data and want to predict their response to drug treatment. Therefore, they first train a $\beta$-TCVAE model to learn a disentangled representation. Then they use the latent space representation of a cell, given by the encoder of the VAE, as additional input for the generator of a conditional GAN. With this approach, they are able to identify latent space dimensions controlling different aspects of cellular identity and predict unseen combinations of cell states.

### Analyze Cancer Genome Atlas

In the work of [Kompa and Coker, 2020], the authors apply a Unified Disentanglement Network (UFDN) [Liu et al., 2014] to the Cancer Genome Atlas (TCGA) [Ramos et al., 2020]. They want to achieve a disentangled latent space representation of the cancer gene expression data to be able to interpolate between cancer types. Therefore, a standard Variational Autoencoder architecture is trained with different types of input data. The network (UFDN) learns to distinguish input domains by a discriminator applied to the latent space of the VAE, while an additional discriminator of the output of the VAE ensures high decoding quality. They applied the model to two classification tasks. The model achieved comparable results like a Random Forest and, up to the authors, additionally offers a biologically relevant latent space representation of the Cancer Genome Atlas data.

### Cell Characterization

In [Bica et al., 2020], the authors investigate the characterization of cells using gene expression data. They introduce DiffVAE, a Variational Autoencoder that can be used to model and analyze the differentiation of cells using single-cell RNA-seq data, and a Graph-DiffVAE. They analyze zebrafish single-cell data [Athanasiadis et al., 2017] and human pancreatic cells [Muraro et al., 2016]. The DiffVAE is an MMD-VAE, this means the Kullback Leibler Divergence in the loss function is replaced with the maximum mean discrepancy (MMD) divergence between $q(z)$ and $p(z)$. Up to the authors, the latent space disentanglement is encouraged by assuming $p(z)$ as a normal distribution with a diagonal covariance matrix. They introduce a pipeline that includes a clustering mechanism of the latent representation of the DiffVAE. They detect important genes for each cluster and are able to map from clusters to cell types, and compare their model to a Variational Autoencoder



(VAE), an Autoencoder (AE), and Principal Component Analysis (PCA) with respect to the cluster performance, where the clustering of the latent space representations of DiffVAE achieved the highest adjusted rand score (ARI).

## Other Medical Applications

### Histopathological Images

For histopathological images, color differences make it challenging to analyze them with DNNs. To address this Moghadam et al. [2022] introduce two GAN-based networks which use disentanglement strategies to separate structural factors which are uniform between data sets and colors which are specific for each data set. The first network is designed for one-to-one tasks while the second one is designed for many-to-many tasks.

### CT Artifact Reduction

Inspired by the success of $\beta$-VAE, Lee et al. [2021] introduced a parameter $\beta$ in a CycleGAN [Zhu et al., 2017], to remove noise resulting from metal artifacts from CT scans, which is known as metal artifact reduction (MAR). They want to disentangle the artifact information from the artifact-free information. They want to learn the distribution of artifact-free images and images with artifacts. Therefore, they apply a generator to transform an artifact-free image to an artifact image and vice versa with another generator. Weighting the first generator more in the loss function, than the second one, disentangles the metal artifact from the generation process and makes the MAR more efficient. Also, in the work of [Liao et al., 2020], the authors consider artifact reduction, by introducing a network denoted as Artifact Disentanglement Network (ADN). They want to disentangle the information about the artifacts, e.g. from hip prostheses, from the content information, e.g. anatomical information, in the latent space representation. Reconstructing Computed Tomography (CT) images without the artifact factor in the latent space, reduces artifact influence on the CT scans. They compare a group of CT scans with artifacts, with a group of CT scans without artifacts. The network consists of several encoders, two encoders get the artifact image as input, one learns the artifact factor representation and the other one learns the content representation. They compare their unsupervised method to CycleGAN [Zhu et al., 2017] and other unsupervised approaches, but also to supervised approaches like a U-Net [Ronneberger et al., 2015]. Among the unsupervised setting, their network achieved higher image quality than the compared baselines.

### Nuclei Segmentation

To analyze three-dimensional cell cultures, one of the basic requirements is to identify individual cells or cell nuclei, which is denoted as nuclei segmentation. To address this challenge the authors of [Yao et al., 2022] introduce a GAN-based network which they denote as Aligned Disentangled Generative Adversarial Network. Within this network, they separate the content representation, which means the spatial structure of the nuclei, from the style representation, which means the rendering of the structure.

### Optical Coherence Tomography Images Noise Reduction

The authors of [Huang et al., 2021] introduce a GAN-based network, called Disentangled Representation and Generative Adversarial Network (DRGAN), to clean and denoise Optical Coherence Tomography (OCT) images. To train the model they need images without noise, which can be obtained by taking multiple image acquisitions from the same position and average over them. For the test prediction, they only need the noisy image. They disentangle the content and noise information by training separate encoder and generator networks because they assume that a noisy image can be separated into noise and content, while a clean image only consists of content. They use a loss function inspired by CycleGAN [Zhu et al., 2017] and compare their results to several GAN-based methods, achieving a high level of noise reduction.

### Protein Self-Assembly

The authors of [Kalinin et al., 2021] analyze the structural development in systems of interacting anisotropic particles. They analyze the particle dynamics during the protein self-assembly process. Therefore, they use an ensemble approach. That means they train several convolutional networks,



here modifications of a U-Net [Ronneberger et al., 2015], in parallel to predict a segmentation map of atomic force microscopy (AFM) data. Then the output of this ensemble is used as the input of a rotationally invariant Variational Autoencoder (rVAE) [Bepler et al., 2019]. In the latent space of the rVAE, they could disentangle the particle rotation from other factors of variation.

**Gait Recognition**

The authors of [Zhang et al., 2020] analyze RGB videos for analyzing human gait. They claim that existing methods either suffer from artifacts like clothing or carried things or have high computational costs. Therefore, they introduce an Autoencoder combined with an Long Short-Term Memory (LSTM), where they want to disentangle the latent space into appearance, canonical, and pose features. They claim that with this disentanglement strategy they outperform state-of-the-art networks.

**Drug Discovery**

The authors Polykovskiy et al. [2018] modified an adversarial Autoencoder [Makhzani et al., 2016] with two supervised disentanglement approaches to generate new molecules with specific properties for drug discovery. Therefore, they want to disentangle the latent space from some properties of the proteins like solubility or simple synthesis possibility. They denote the first disentanglement approach as predictive disentanglement, where they minimize the mutual information between the properties and the latent space. The second approach is denoted as joint disentanglement where a discriminator learns to separate pairs of the latent representation with the properties from pairs from noise. For the final model, they combine these two approaches and are able to generate novel molecules with favorable scaffolds.

# 6 Discussion

Disentanglement approaches offer a great potential to overcome the black box characteristics of DNNs, important for reliable medical applications. But while being promising, several challenges remain to be addressed:

**Real-World Applications**

With the results from our review with the search string *Disentanglement[Title/Abstract]* restricted to the period 01/01/2017 to 01/13/2022 on PubMed, we showed that the concept of disentangled representations is widespread. Several methods applied disentanglement strategies to improve the performance of their networks or to gain control about different factors of variation. A huge range of data modalities is already considered. Imaging data are most strongly represented, led by MR and CT scans. But also applications to PET, OCT, and whole slide images are present, as well as several other data modalities like ECG or Cell-data. In the reviewed studies, one of the most popular disentanglement strategies is to include a priori knowledge about the data into the training process and separate the latent space representation into two parts, like into content and style. But also methods without this bias, like in [Higgins et al., 2021], found their way to medical context. Furthermore, an often applied use case in the reviewed medical applications is to use the disentangled latent space representation for regression tasks to predict a factor of interest. But also other applications important for medical context, like image denoising, missing modality handling and improving classification tasks through synthetic data usage benefit from the disentanglement idea. Finally, to the best of our knowledge, some of the technical disentanglement approaches have not yet been applied to medical data, e.g. Hessian Penalty [Peebles et al., 2020] or [Lin et al., 2020], thus, there is potential for further studies. Future challenges will still be to disentangle latent representations with as little a priori knowledge of the data as possible while still being identifiable.

**Generalization**

The authors of [Montero et al., 2021] analyzed three classes of Variational Autoencoders (VAEs) for their generalization performance. Up to them, disentangled representations increases interpretability of a network and improve sample efficiency in some downstreaming tasks but do not boost generalization performance. They assume that disentangled representations are necessary, but not sufficient for the generalization task. Thus, future research could explore if disentanglement combined with other biases



can improve the generalization performance and other generative models, like GANs and Flow-based models, with a disentanglement approach could be analysed for their generalization performance. Furthermore, [Montero et al., 2021] only tested on synthetic data sets like dSprites and 3D shapes, thus, even if generalization performance could be improved there, generalization performance for medical data will probably be further challenging.

**Independence and Compactness**

Different realizations of ground truth factor can exist [Duan et al., 2019]. The request for compact representations may offer a smaller set of possible solutions but may not be the best representation. In [Ridgeway and Mozer, 2018] they note the example that an angle $\theta \in [0, 360]$, describing the orientation of an object, could be represented by $sin(\theta)$ and $cos(\theta)$ reflecting the natural characteristics of the orientation while not being compact. They also argue that a compact representation could lead to local optima while training up to a constrained solution space while allowing more dimensions in the latent code for one factor would offer different equivalently good optima. Interestingly, Higgins et al. [2021] observed that the corespondents of a subset of neurons to the full amount of neurons measured with the Disentanglement score of Eastwood and Williams [2018], was not one, which means that more neurons correspond to single information. Therefore, learning independently distributed dimensions may not be the way a human brain learns and latent space representations should be organized. For some medical applications one could ask if independence between the latent dimensions is necessary. Imagine a representation with the factors of variation: *Gender, Chest size and Pelvis size*. If gender would be female, the latent space dimensions representing chest size and pelvis size follow a different distribution, as if the gender would be male or divers. Thus, these dimensions would not be independent, but this representation would offer more control to generate realistic images. Hence, future research should concentrate on a uniform definition that is clearly describing the necessary and requested properties of disentangled latent representations.

**Model-agnostic Approaches**

In [Lin et al., 2020] they tried the *FactorVAE* [Kim and Mnih, 2019] regularizer on their InfoGAN-CR approach and compare it to their model without this regularizer, which obtained a better disentanglement score, with respect to the FactorVAE metric on the dSprites dataset. On the other hand, they trained a FactorVAE on dSprites and for the last batches they add the CR-regularizer, but the disentanglement score did not improve. Thus, they hypothesize that GAN and VAE models need different disentangling approaches, doubting the idea of a model-agnostic disentanglement approach. Further research could explore if different architectures really need different disentanglement strategies, and if so, which strategy works best for which network, or if one fundamental disentanglement paradigm exists. Such an paradigm would revolutionize this field, but requires a deep understanding of the theoretical background.

**Evaluation**

The evaluation of a disentangled representation is a challenging task. Most metrics need ground truth factors to be computed. The possibility of different realizations of ground truth factor affects these measurements [Duan et al., 2019] and, hence, supervised metrics may not catch the real degree of a disentangled representation. Apart from this, nearly all metrics contain some failure modes, as [Sepliarskaia et al., 2021] showed, thus, results should be used carefully. And also the need of hyperparameter optimization or the architecture-dependence of some metrics make results hard to compare. Therefore, future challenges will be the development of model-independent, unsupervised and hyperparameter-free metrics. Especially in healthcare, the focus should be on unsupervised ways for evaluation, as ground truth information is often not available.

# 7 Conclusion

We presented several technical approaches for disentanglement networks and review medical applications found on PubMed. According to our search string we mainly found VAE- and GAN-based models applied for medical purposes, but no Flow-based



models. The most disentanglement approaches tried to separate content and style and boost network performance or improved interpretability of the latent space, helping to overcome the black box characteristic of the networks. To conclude, we

- provided an overview of current medical imaging works, where disentanglement plays the key role;
- conducted a comprehensive introduction and background foundation of disentanglement;
- introduced common metrics used in disentanglement;
- extracted challenges and future work in the field of medical disentanglement.

**Acknowledgement**

We acknowledge the REACT-EU project KITE (Plattform für KI-Translation Essen).

# References


Abbasi, A., Monadjemi, A., Fang, L., Rabbani, H., 2018. Optical coherence tomography retinal image reconstruction via nonlocal weighted sparse representation. Journal of Biomedical Optics 23, 1. doi:10.1117/1.JBO.23.3.036011.

Arjovsky, M., Chintala, S., Bottou, L., 2017. Wasserstein generative adversarial networks, in: International conference on machine learning, PMLR. pp. 214–223.

Armato III, S.G., McLennan, G., Bidaut, L., McNitt-Gray, M.F., et. al., 2011. The lung image database consortium (lidc) and image database resource initiative (idri): a completed reference database of lung nodules on ct scans. Medical Physics, 38(2):915–931 .

Athanasiadis, E., et al., 2017. Single-cell rna-sequencing uncovers transcriptional states and fate decisions in haematopoiesis. Nature communications 8, 2045, https://doi.org/10.1038/s41467-017-02305-6.

Bastidas-Ponce, A., Tritschler, S., Dony, L., Scheibner, K., Tarquis-Medina, M., Salinno, C., Schirge, S., Burtscher, I., Böttcher, A., Theis, F.J., Lickert, H., Bakhti, M., Klein, A., Treutlein, B., 2019. Comprehensive single cell mRNA profiling reveals a detailed roadmap for pancreatic endocrinogenesis. Development 146. URL: https://doi.org/10.1242/dev.173849, doi:10.1242/dev.173849. dev173849.

Behrmann, J., Grathwohl, W., Chen, R.T.Q., Duvenaud, D., Jacobsen, J., 2019. Invertible residual networks, in: Proceedings of the 36th International Conference on Machine Learning, ICML 2019, 9-15 June 2019, Long Beach, California, USA, pp. 573–582, http://proceedings.mlr.press/v97/behrmann19a.html.

Bejnordi, B.E., Veta, M., Van Diest, P.J., Van Ginneken, B., Karssemeijer, N., Litjens, G., Van Der Laak, J.A., Hermsen, M., Manson, Q.F., Balkenhol, M., et al., 2017. Diagnostic assessment of deep learning algorithms for detection of lymph node metastases in women with breast cancer. Jama 318, 2199–2210.

Ben-Cohen, A., Mechrez, R., Yedidia, N., Greenspan, H., 2019. Improving cnn training using disentanglement for liver lesion classification in ct. in: Proc. IEEE Engineering in Medicine and Biology Society (EMBC), pp. 886–889.

Bepler, T., Zhong, E., Kelley, K., Brignole, E., Berger, B., 2019. Explicitly disentangling image content from translation and rotation with spatial-vae. Advances in Neural Information Processing Systems 2019, 15409-15419.

Bica, I., Andrés-Terré, H., Cvejic, A., et al., 2020. Unsupervised generative and graph representation learning for modelling cell differentiation. Sci Rep 10, 9790. https://doi.org/10.1038/s41598-020-66166-8.

Bilic, P., Christ, P., Vorontsov, E., Chlebus, G., Chen, H., Dou, Q., Fu, C.W., Han, X., Heng, P.A., Hesser, J., et al., 2019. The liver tumor segmentation benchmark (lits). arXiv:1901.04056.

Bousquet, O., Gelly, S., Tolstikhin, I., Simon-Gabriel, C.J., Schoelkopf, B., 2017. From optimal transport to generative modeling: the vegan cookbook. arXiv:1705.07642 .





Burgess, C., Kim, H., 2018. 3d shapes dataset. https://github.com/deepmind/3dshapes-dataset/.

Chen, J., Batmanghelich, K., 2020. Weakly supervised disentanglement by pairwise similarities. Proc Conf AAAI Artif Intell. 2020 Feb;34(4). doi:10.1609/aaai.v34i04.5754.

Chen, R.T., Rubanova, Y., Bettencourt, J., Duvenaud, D.K., 2018a. Neural ordinary differential equations. Advances in neural information processing systems 31.

Chen, T.Q., Li, X., Grosse, R.B., Duvenaud, D., 2018b. Isolating sources of disentanglement in variational autoencoders. CoRR abs/1802.04942. URL: http://arxiv.org/abs/1802.04942, arXiv:1802.04942.

Chen, X., Duan, Y., Houthooft, R., Schulman, J., Sutskever, I., Abbeel, P., 2016. Infogan: Interpretable representation learning by information maximizing generative adversarial nets. arXiv:1606.03657.

Christ, P., Ettlinger, F., Grün, F., Lipkova, J., Kaissis, G., 2017. Lits-liver tumor segmentation challenge. ISBI and MICCAI.

Consortium, T.M., et al., 2018. Single-cell transcriptomics of 20 mouse organs creates a tabula muris. Nature. 2018; 562(7727):367-72.

Dinh, L., Krueger, D., Bengio, Y., 2015. Nice: Non-linear independent components estimation. arXiv:1410.8516.

Dinh, L., Sohl-Dickstein, J., Bengio, S., 2016. Density estimation using real nvp. arXiv:1605.08803.

Do, K., Tran, T., 2021. Theory and evaluatin metrics for learning disentangled representations. arXiv:1908.09961.

Donahue, J., Simonyan, K., 2019. Large scale adversarial representation learning, in: Wallach, H., Larochelle, H., Beygelzimer, A., d'Alché-Buc, F., Fox, E., Garnett, R. (Eds.), Advances in Neural Information Processing Systems, Curran Associates, Inc., https://proceedings.neurips.cc/paper/2019/file/18cdf49ea54eec029238fcc95f76ce41-Paper.pdf.

Duan, S., Matthey, L., Saraiva, A., Watters, N., Burgess, C., Lerchner, A., Higgins, I., 2019. Unsupervised model selection for variational disentangled representation learning, in: International Conference on Learning Representations.

Eastwood, C., Williams, C.K.I., 2018. A framework for the quantitative evaluation of disentangled representations, in: International Conference on Learning Representations, https://openreview.net/forum?id=By-7dz-AZ.

Egger, J., Pepe, A., Gsaxner, C., Jin, Y., Li, J., Kern, R., 2021. Deep learning—a first meta-survey of selected reviews across scientific disciplines, their commonalities, challenges and research impact. PeerJ Computer Science 7, e773.

Esser, P., Rombach, R., Ommer, B., 2020. A disentangling invertible interpretation network for explaining latent representations. arXiv:2004.13166.

Fei, Y., Zhan, B., Hong, M., Wu, X., Zhou, J., Wang, Y., 2021. Deep learning-based multi-modal computing with feature disentanglement for mri image synthesis. Med Phys. 2021 Jul;48(7):3778-3789. doi: 10.1002/mp.14929. Epub 2021 Jun 7. PMID: 33959965.

Gao, W., et al., 2008. The cas-peal large-scale chinese face database and baseline evaluations. .

Gillies, R.J., Kinahan, P.E., Hricak, H., 2016. Radiomics: Images are more than pictures, they are data. Radiology 278, 563–577. URL: https://doi.org/10.1148/radiol.2015151169, doi:10.1148/radiol.2015151169, arXiv:https://doi.org/10.1148/radiol.2015151169. pMID: 26579733.

Goodfellow, I.J., Pouget-Abadie, J., Mirza, M., Xu, B., Warde-Farley, D., Ozair, S., Courville, A., Bengio, Y., 2014. Generative adversarial networks. arXiv:1406.2661.

Grathwohl, W., Chen, R.T., Bettencourt, J., Sutskever, I., Duvenaud, D., 2018. Ffjord: Free-form continuous dynamics for scalable reversible generative models, in: International Conference on Learning Representations.





Griffiths, T.L., Ghahramani, Z., 2011. The indian buffet process: An introduction and review. Journal of Machine Learning Research 12, 1185–1224, http://jmlr.org/papers/v12/griffiths11a.html.

Gyawali, P., Murkute, J., Toloubidokhti, M., Jiang, X., Horacek, B., Sapp, J., Wang, L., 2021. Learning to disentangle inter-subject anatomical variations in electrocardiographic data. IEEE Trans Biomed Eng. 2021 Aug 30;PP. doi: 10.1109/TBME.2021.3108164. Epub ahead of print. PMID: 34460360.

Gyawali, P.K., Horacek, B.M., Sapp, J.L., Wang, L., 2019. Sequential factorized autoencoder for localizing the origin of ventricular activation from 12-lead electrocardiograms. IEEE Transactions on Biomedical Engineering 67, 1505–1516.

Havaei, M., Mao, X., Wang, Y., Lao, Q., 2021. Conditional generation of medical images via disentangled adversarial inference. Medical Image Analysis 72, 102106.

Higgins, I., Amos, D., Pfau, D., Racaniere, S., Matthey, L., Rezende, D., Lerchner, A., 2018. Towards a definition of disentangled representations. arXiv:1812.02230.

Higgins, I., Chang, L., Langston, V., Hassabis, D., Summerfield, C., Tsao, D., Botvinick, M., 2021. Unsupervised deep learning identifies semantic disentanglement in single inferotemporal face patch neurons. Nature communications 12, 1–14.

Higgins, I., Matthey, L., Pal, A., Burgess, C.P., Glorot, X., Botvinick, M., Mohamed, S., Lerchner, A., 2017. beta-vae: Learning basic visual concepts with a constrained variational framework, in: ICLR.

Howell, B., et al., 2019. "the unc/umn baby connectome project (bcp): an overview of the study design and protocol development,". NeuroImage, vol. 185, pp. 891–905, 2019. [PubMed: 29578031] .

Hu, D., Wang, F., Zhang, H., Wu, Z., Wang, L., Lin, W., Li, G., Shen, D., Consortium, U.B.C.P., et al., 2020. Disentangled intensive triplet autoencoder for infant functional connectome fingerprinting, in: International Conference on Medical Image Computing and Computer-Assisted Intervention, Springer. pp. 72–82.

Hu, D., Zhang, H., Wu, Z., Wang, F., Wang, L., Smith, J.K., Lin, W., Li, G., Shen, D., 2021. Disentangled-multimodal adversarial autoencoder: Application to infant age prediction with incomplete multimodal neuroimages. IEEE Trans Med Imaging. 2020 Dec;39(12):4137-4149. doi: 10.1109/TMI.2020.3013825. Epub 2020 Nov 30. PMID: 32746154; PMCID: PMC7773223.

Huang, Y., Xia, W., Lu, Z., Liu, Y., Chen, H., Zhou, J., Fang, L., Zhang, Y., 2021. Noise-powered disentangled representation for unsupervised speckle reduction of optical coherence tomography images. IEEE Trans Med Imaging. 2021 Oct;40(10):2600-2614. doi: 10.1109/TMI.2020.3045207. Epub 2021 Sep 30. PMID: 33326376.

Hyvärinen, A., Oja, E., 2000. Independent component analysis: algorithms and applications. Neural Netw. 2000 May-Jun;13(4-5):411-30. doi:doi:10.1016/s0893-6080(00)00026-5.PMID:10946390.

Justusson, B.I., 1981. Median Filtering: Statistical Properties. Springer Berlin Heidelberg, Berlin, Heidelberg. pp. 161–196. URL: https://doi.org/10.1007/BFb0057597, doi:10.1007/BFb0057597.

Jutten, C., Herault, J., 1991. Blind separation of sources, part i: An adaptive algorithm based on neuromimetic architecture. Signal Processing 24, 1–10. URL: https://www.sciencedirect.com/science/article/pii/016516849190079X, doi:https://doi.org/10.1016/0165-1684(91)90079-X.

Kalinin, S.V., Zhang, S., Valleti, M., Pyles, H., Baker, D., Yoreo, J.J.D., Ziatdinov, M., 2021. Disentangling rotational dynamics and ordering transitions in a system of self-organizing protein nanorods via rotationally invariant latent representations. ACS Nano. 2021 Apr 27;15(4):6471-6480. doi: 10.1021/acsnano.0c08914. Epub 2021 Apr 16. PMID: 33861068.

Karras, T., Laine, S., Aila, T., 2019. A style-based generator architecture for generative adversarial networks, in: Proceedings of the IEEE/CVF conference on computer vision and pattern recognition, pp. 4401–4410.





Karras, T., Laine, S., Aittala, M., Hellsten, J., Lehtinen, J., Aila, T., 2020. Analyzing and improving the image quality of stylegan. arXiv:1912.04958.

Khemakhem, I., Kingma, D.P., Monti, R.P., Hyvärinen, A., 2020. Variational autoencoders and nonlinear ica: A unifying framework. arXiv:1907.04809.

Kim, H., Mnih, A., 2019. Disentangling by factorising. arXiv:1802.05983.

Kim, M., Wang, Y., Sahu, P., Pavlovic, V., 2019. Relevance factor vae: Learning and identifying disentangled factors. arXiv:1902.01568.

Kingma, D.P., Dhariwal, P., 2018. Glow: Generative flow with invertible 1x1 convolutions. arXiv:1807.03039.

Kingma, D.P., Welling, M., 2014. Auto-encoding variational bayes. arXiv:1312.6114.

Kleesiek, J., Kersjes, B., Ueltzhöffer, K., Murray, J.M., Rother, C., Köthe, U., Schlemmer, H.P., 2021. Discovering digital tumor signatures—using latent code representations to manipulate and classify liver lesions. Cancers 13. URL: https://www.mdpi.com/2072-6694/13/13/3108, doi:10.3390/cancers13133108.

Kompa, B., Coker, B., 2020. Learning a latent space of highly multidimensional cancer data. Pacific Symposium on Biocomputing. Pacific Symposium on Biocomputing vol. 25 (2020): 379-390.

Kromp, F., Bozsaky, E., Rifatbegovic, F., Fischer, L., Ambros, M., Berneder, M., Weiss, T., Lazic, D., Dörr, W., Hanbury, A., et al., 2020. An annotated fluorescence image dataset for training nuclear segmentation methods. Scientific Data 7, 1–8.

Kumar, A., Sattigeri, P., Balakrishnan, A., 2018. Variational inference of disentangled latent concepts from unlabeled observations, in: International Conference on Learning Representations.

Lang, O., Gandelsman, Y., Yarom, M., Wald, Y., Elidan, G., Hassidim, A., Freeman, W.T., Isola, P., Globerson, A., Irani, M., Mosseri, I., 2021. Explaining in style: Training a gan to explain a classifier in stylespace. arXiv:2104.13369.

Lee, J., Gu, J., Ye, J.C., 2021. Unsupervised ct metal artifact learning using attention-guided -cyclegan. IEEE Transactions on Medical Imaging 40, 3932–3944. doi:10.1109/TMI.2021.3101363.

Li, J., Pimentel, P., Szengel, A., Ehlke, M., Lamecker, H., Zachow, S., Estacio, L., Doenitz, C., Ramm, H., Shi, H., et al., 2021. Autoimplant 2020-first miccai challenge on automatic cranial implant design. IEEE transactions on medical imaging 40, 2329–2342.

Li, J., Yang, S., Huang, X., Da, Q., Yang, X., Hu, Z., Duan, Q., Wang, C., Li, H., 2019. Signet ring cell detection with a semi-supervised learning framework, in: International conference on information processing in medical imaging, Springer. pp. 842–854.

Liao, H., Lin, W.A., Zhou, S.K., Luo, J., 2020. Adn: Artifact disentanglement network for unsupervised metal artifact reduction. IEEE Trans Med Imaging. 2020 Mar;39(3):634-643. doi: 10.1109/TMI.2019.2933425. Epub 2019 Aug 5. PMID: 31395543.

Lin, Z., Thekumparampil, K.K., Fanti, G., Oh, S., 2020. Infogan-cr and modelcentrality: Self-supervised model training and selection for disentangling gans. arXiv:1906.06034.

Litjens, G., Debats, O., Barentsz, J., Karssemeijer, N., Huisman, H., 2014. Computer- aided detection of prostate cancer in mri.

Liu, A., Liu, Y.C., Yeh, Y.Y., Wang, Y.C., 2014. A unified feature disentangler for multi-domain image translation and manipulation. in Advances in Neural Information Processing Systems 31, eds. Bengio S, Wallach H, Larochelle H, Grauman K, Cesa-Bianchi N and Garnett R (Curran Associates, Inc., 2018) pp. 2595–2604.

Liu, X., Sanchez, P., Thermos, S., O'Neil, A.Q., Tsaftaris, S.A., 2021a. Learning disentangled representations in the imaging domain. arXiv:2108.12043.

Liu, X., Xing, F., Fakhri, G.E., Woo, J., 2021b. A unified conditional disentanglement framework for multimodal brain mr image translation. Proc IEEE Int Symp Biomed Imaging. 2021





Apr;2021:10.1109/isbi48211.2021.9433897. doi: 10.1109/isbi48211.2021.9433897. PMID: 34567419; PMCID: PMC8460116.

Liu, Z., Luo, P., Wang, X., Tang, X., 2015. Deep learning face attributes in the wild. .

Locatello, F., Bauer, S., Lucic, M., Gelly, S., Schölkopf, B., Bachem, O., 2018. Challenging common assumptions in the unsupervised learning of disentangled representations. CoRR abs/1811.12359. URL: http://arxiv.org/abs/1811.12359, arXiv:1811.12359.

Locatello, F., Bauer, S., Lucic, M., Raetsch, G., Gelly, S., Schölkopf, B., Bachem, O., 2020. A sober look at the unsupervised learning of disentangled representations and their evaluation. Journal of Machine Learning Research 21, 1–62. URL: http://jmlr.org/papers/v21/19-976.html.

Ma, D.S., Correll, J., Wittenbrink, B., 2015. The chicago face database: a free stimulus set of faces and norming data. .

Maier, O., Menze, B., von der Gablentz, J., H ani, L and, H.M., Liebrand, M., Winzeck, S., Basit, A., Bentley, P., Chen, L., et al., 2015. Isles 2015-a public evaluation benchmark for ischemic stroke lesion segmentation from multispectral mri. Medical image analysis 35, 250–269. .

Makhzani, A., 2019. Implicit autoencoders. arXiv:1805.09804.

Makhzani, A., Shlens, J., Jaitly, N., Goodfellow, I., Frey, B., 2016. Adversarial autoencoders. arXiv:1511.05644.

Mathieu, E., Rainforth, T., Siddharth, N., Teh, Y.W., 2019. Disentangling disentanglement in variational autoencoders. arXiv:1812.02833.

Matthey, L., Higgins, I., Hassabis, D., Lerchner, A., 2017. dsprites: Disentanglement testing sprites dataset. https://github.com/deepmind/dsprites-dataset/.

Menze, B., et al., 2014. The multimodal brain tumor image segmentation benchmark (brats). IEEE transactions on medical imaging 34, 1993–2024.

Menze, B., et al., 2015. The multimodal brain tumor image segmentation benchmark (brats). IEEE transactions on medical imaging 34, 1993–2024.

Mitchell, G.F., Hwang, S.J., Vasan, R.S., Larson, M.G., Pencina, M.J., Hamburg, N.M., Vita, J.A., Levy, D., Benjamin, E.J., 2010. Arterial stiffness and cardiovascular events: the framingham heart study. Circulation 121(4): 505.

Moghadam, A.Z., Azarnoush, H., Seyyedsalehi, S.A., Havaei, M., 2022. Stain transfer using generative adversarial networks and disentangled features. Computers in Biology and Medicine , 105219.

Moher, D., Liberati, A., Tetzlaff, J., Altman, D.G., Group, P., et al., 2009. Preferred reporting items for systematic reviews and meta-analyses: the prisma statement. PLoS medicine 6, e1000097.

Montero, M.L., Ludwig, C.J., Costa, R.P., Malhotra, G., Bowers, J., 2021. The role of disentanglement in generalisation, in: International Conference on Learning Representations. URL: https://openreview.net/forum?id=qbH974jKUVy.

Moody, G.B., Mark, R.G., 2001. The impact of the mitbih arrhythmia database. IEE Engineering in Medicine and Biology Magazine, vol.20, no. 3, pp. 45-50.

Muraro, M.J., et al., 2016. A single-cell transcriptome atlas of the human pancreas. Cell systems 3, 385–394 (2016).

Nensa, F., Demircioglu, A., Rischpler, C., 2019. Artificial intelligence in nuclear medicine. Journal of Nuclear Medicine 60, 29S–37S. URL: https://jnm.snmjournals.org/content/60/Supplement_2/29S, doi:10.2967/jnumed.118.220590.

Ouyang, J., Adeli, E., Pohl, K.M., Zhao, Q., Zaharchuk, G., 2021. Representation disentanglement for multi-modal brain mri analysis, in: International Conference on Information Processing in Medical Imaging, Springer. pp. 321–333.

Peebles, W., Peebles, J., Zhu, J.Y., Efros, A., Torralba, A., 2020. The hessian penalty: A weak prior for unsupervised disentanglement. arXiv:2008.10599.

Peer, P., 1999. Cvl face database .





Phillips, P., Wechsler, H., Huang, J., Rauss, P., 1998. The feret database and evaluation procedure for face recognition algorithms. .

Polykovskiy, D., Alexander Zhebrak, D.V., Ivanenkov, Y., Aladinskiy, V., Mamoshina, P., Bozdaganyan, M., Aliper, A., Zhavoronkov, A., Kadurin, A., 2018. Entangled conditional adversarial autoencoder for de novo drug discovery. Mol Pharm. 2018 Oct 1;15(10):4398-4405. doi: 10.1021/acs.molpharmaceut.8b00839. Epub 2018 Sep 19. PMID: 30180591.

Quan, S.F., V Howard, B., Iber, C., Kiley, J., Nieto, F., O'Connor, G., Rapoport, D., Redline, S., Robbins, J., Samet, J.e.a., 1997. "the sleep heart health study: design, rationale, and methods.". Sleep 20(12):1077–1085. .

Ramos, M., Waldron, L., Schiffer, L., Geistlinger, L., Obenchain, V., Morgan, M., 2020. curatedtcga data: Curated data from the cancer genome atlas (tcga) as multiassayexperiment objects. R package version 1.

Ren, X., Yang, T., Wang, Y., Zeng, W., 2021. Do generative models know disentanglement? contrastive learning is all you need. arXiv:2102.10543.

Ridgeway, K., Mozer, M.C., 2018. Learning deep disentangled embeddings with the f-statistic loss. arXiv:1802.05312.

Ries, L.G., Young, J., Keel, G., Eisner, M., Lin, Y., Horner, M., et al., 2007. Seer survival monograph: cancer survival among adults: Us seer program, 1988-2001, patient and tumor characteristics. National Cancer Institute, SEER Pro- gram, NIH Pub (07-6215): 193–202. .

Robinson, C., Trivedi, A., Blazes, M., Ortiz, A., Desbiens, J., Gupta, S., Dodhia, R., Bhatraju, P.K., Liles, W.C., Lee, A., et al., 2021. Deep learning models for covid-19 chest x-ray classification: Preventing shortcut learning using feature disentanglement. medRxiv .

Ronneberger, O., Fischer, P., Brox, T., 2015. U-net: Convolutional networks for biomedical image segmentation. arXiv:1505.04597.

Roux, L., Racoceanu, D., Capron, F., Calvo, J., Attieh, E., Le Naour, G., Gloaguen, A., 2014. Mitos & atypia. Image Pervasive Access Lab (IPAL), Agency Sci., Technol. & Res. Inst. Infocom Res., Singapore, Tech. Rep 1, 1–8.

Sankar, A., Keicher, M., Eisawy, R., Parida, A., Pfister, F., Kim, S.T., Navab, N., 2021. Glowin: A flow-based invertible generative framework for learning disentangled feature representations in medical images. arXiv:2103.10868.

Sarkar, S., Phillips, P.J., Liu, Z., Vega, I.R., Grother, P., Bowyer, K.W., 2005. "the human id gait challenge problem: Data sets, performance, and analysis,". IEEE Transactions on Pattern Analysis and Machine Intelligence (TPAMI), vol. 27, no. 2, pp. 162–177, 2005.

Sepliarskaia, A., Kiseleva, J., de Rijke, M., 2021. How to not measure disentanglement. arXiv:1910.05587.

Shen, L., Zhu, W., Wang, X., Xing, L., Pauly, J.M., Turkbey, B., Harmon, S.A., Sanford, T.H., Mehralivand, S., Choyke, P.L., Wood, B.J., Xu, D., 2021. Multi-domain image completion for random missing input data. IEEE Trans Med Imaging. 2021 Apr;40(4):1113-1122. doi: 10.1109/TMI.2020.3046444. Epub 2021 Apr 1. PMID: 33351753; PMCID: PMC8136445.

Shu, R., Chen, Y., Kumar, A., Ermon, S., Poole, B., 2020. Weakly supervised disentanglement with guarantees. arXiv:1910.09772.

Simonyan, K., Zisserman, A., 2014. Very deep convolutional networks for large-scale image recognition. arXiv preprint arXiv:1409.1556 .

Simonyan, K., Zisserman, A., 2015. Very deep convolutional networks for large-scale image recognition. arXiv:1409.1556.

Song, J., Shi, J., Dong, D., et al., 2018. A new approach to predict progression-free survival in stage iv egfr-mutant nsclc patients with egfr-tki therapy. Clin Cancer Res. 2018;24(15):3583-3592. doi:10.1158/1078-0432.CCR-17-2507.

Song, J., Wang, L., Ng, N.N., Zhao, M., Shi, J., Ning Wu, W.L., Liu, Z., Yeom, K.W., Tian, J., 2020. Development and validation of a machine learning model





to explore tyrosine kinase inhibitor response in patients with stage iv egfr variant-positive non-small cell lung cancer. JAMA Netw Open. 2020 Dec 1;3(12):e2030442. doi: 10.1001/jamanetworkopen.2020.30442. Erratum in: JAMA Netw Open. 2021 Feb 1;4(2):e211634. PMID: 33331920; PMCID: PMC7747022. .

Sorrenson, P., Rother, C., Köthe, U., 2020. Disentanglement by nonlinear ica with general incompressible-flow networks (gin). arXiv:2001.04872.

Srivatsan, S., McFaline-Figueroa, J., Ramani, V., Saunders, L., Cao, J., Packer, J., Pliner, H., Jackson, D., Daza, R., Christiansen, L., et al., 2020. Massively multiplex chemical transcriptomics at single-cell resolution. Science. 2020;367(6473): 45–51.

Strohminger, N., et al., 2016. The mr2: a multi-racial mega-resolution database of facial stimuli. .

Taylor, J., Williams, N., Cusack, R., Auer, T., Shafto, M., Dixon, M., Tyler, L., Henson, R., et al., 2017. The cambridge centre for ageing and neuroscience (cam-can) data repository: structural and functional mri, meg, and cognitive data from a cross-sectional adult lifespan sample. Neuroimage 144, 262–269.

Toda, R., Teramoto, A., Tsujimoto, M., et al., 2021. Synthetic ct image generation of shape-controlled lung cancer using semi-conditional infogan and its applicability for type classification. Int J CARS 16, 241–251 (2021).s , 101719doi:https://doi.org/10.1007/s11548-021-02308-1.

Tschandl, P., Rosendahl, C., Kittler., H., 2018. The ham10000 dataset, a large collection of multi-source dermatoscopic images of common pigmented skin lesions. Scientific Data, 5, 2018. .

Van Steenkiste, T., Deschrijver, D., Dhaene, T., 2019. Interpretable ecg beat embedding using disentangled variational auto-encoders. Medical Image Analysis 32, 101719. doi:10.1109/CBMS.2019.00081.

Velten, L., et al., 2017. Human haematopoietic stem cell lineage commitment is a continuous process. Nature cell biology 19, 271 (2017).

Wang, T., Lei, Y., Fu, Y., Curran, W.J., Liu, T., Yang, X., 2020. Medical imaging synthesis using deep learning and its clinical applications: A review. arXiv:2004.10322.

Weinstein, J.N., Collisson, E.A., Mills, G.B., Shaw, K.R., Ozenberger, B.A., Ellrott, K., Shmulevich, I., Sander, C., Stuart, J.M., 2013. The cancer genome atlas pan-cancer analysis project. Nature genetics 45, 1113–1120.

Willetts, M., Paige, B., 2021. I don't need **u**: Identifiable non-linear ica without side information. arXiv:2106.05238.

Xia, T., Chartsias, A., Tsaftaris, S.A., 2020. Pseudo-healthy synthesis with pathology disentanglement and adversarial learning. Medical Image Analysis 64, 101719. URL: http://dx.doi.org/10.1016/j.media.2020.101719, doi:10.1016/j.media.2020.101719.

Xiu, Z., Tao, C., Gao, M., Davis, C., Goldstein, B.A., Henao, R., 2021. Variational disentanglement for rare event modeling. arXiv:2009.08541.

Yan, K., Wang, X., Lu, L., Summers, R.M., 2018. Deeplesion: Automated mining of large-scale lesion annotations and universal lesion detection with deep learning. J. Med. Imag., vol. 5, no. 3, Jul. 2018, Art. no. 036501.

Yang, J., Dvornek, N.C., Zhang, F., Zhuang, J., Chapiro, J., Lin, M., Duncan, J.S., 2019. Domain-agnostic learning with anatomy-consistent embedding for cross-modality liver segmentation, in: Proceedings of the IEEE/CVF International Conference on Computer Vision Workshops, pp. 0–0.

Yao, K., Sun, J., Huang, K., Jing, L., Liu, H., Huang, D., Jude, C., 2022. Analyzing cell-scaffold interaction through unsupervised 3d nuclei segmentation. International journal of bioprinting 8.

Yu, H., Welch, J., 2021. Michigan: sampling from disentangled representations of single-cell data using generative adversarial networks. Genome Biol. 2021 May 20;22(1):158. doi:doi:10.1186/s13059-021-02373-4.





Yu, S., Tan, D., Tan, T., 2006. A framework for evaluating the effect of view angle, clothing and carrying condition on gait recognition. in International Conference on Pattern Recognition (ICPR).

Zaidi, J., Boilard, J., Gagnon, G., Carbonneau, M.A., 2020. Measuring disentanglement: A review of metrics. arXiv:2012.09276.

Zech, J.R., Badgeley, M.A., Manway Liu, A.B.C., Titano, J.J., Oermann, E.K., 2018. Variable generalization performance of a deep learning model to detect pneumonia in chest radiographs: A cross-sectional study. Neural Netw. 2000 May-Jun;13(4-5):411-30. doi:PLoSMed.2018Nov6;15(11): e1002683.doi:10.1371/journal.pmed.1002683. PMID:30399157;PMCID:PMC6219764.

Zhang, Z., Tran, L., Liu, F., Liu, X., 2020. On learning disentangled representations for gait recognition. IEEE Transactions on Pattern Analysis and Machine Intelligence .

Zhao, Q., Adeli, E., Honnorat, N., Leng, T., Pohl, K.M., 2019. Variational autoencoder for regression: Application to brain aging analysis, in: International Conference on Medical Image Computing and Computer-Assisted Intervention, Springer. pp. 823–831.

Zhao, Q., Liu, Z., Adeli, E., Pohl, K.M., 2021. Longitudinal self-supervised learning. Medical Image Analysis 71, 102051.

Zhao, Q., Sullivan, E., Honnorat, N., Adeli, E., Podhajsky, S., De Bellis, M., Voyvodic, J., Nooner, K., Baker, F., Colrain, I., et al., 2020. Association of heavy drinking with deviant fiber tract development in frontal brain systems in adolescents. JAMA Psychiatry .

Zhou, S., Zelikman, E., Lu, F., Ng, A.Y., Carlsson, G.E., Ermon, S., 2021. Evaluating the disentanglement of deep generative models through manifold topology, in: International Conference on Learning Representations, https://openreview.net/forum?id=djwS0m4Ft_A.

Zhu, J.Y., Park, T., Isola, P., Efros, A., 2017. Unpaired image-to-image translation using cycle consistent adversarial networks. In Proceedings of the IEEE international conference on computer vision, pages 2223–2232.


# Appendix:

## 7.1 Information theory

**Entropy**
The Entropy of a discrete random variable $X$ is defined by

$$H(X) = \sum_{x \in \mathcal{X}} p(x) \log(p(x))$$

where $p$ is the probability mass function. It can be analogously defined for continuous random variables with a density function $p$:

$$H(X) = \int_{\mathcal{X}} p(x) \log(p(x)) dx$$

The Entropy of a random variable $X$ measures its uncertainty. This means, equally distributed random variables have a high entropy as every state is equally likely. Otherwise, a random variable like:

$$X = \begin{cases} 1 & \text{with probability } 0.99 \\ 0 & \text{with probability } 0.01 \end{cases}$$

has a low entropy as it is kind of reliable that the value of the random variable will be one.

**Mutual Information:**
For two discrete random variables $X$ and $Y$, jointly distributed with respect to the joint probability mass function $p(x,y)$ the mutual information is given as:

$$I(X,Y) = \sum_{x \in \mathcal{X}} \sum_{y \in \mathcal{Y}} p(x,y) \log\left(\frac{p(x,y)}{p(y)p(x)}\right)$$

This can analogously be defined for continuous variables with density function:

$$I(X,Y) = \int_{\mathcal{X}} \int_{\mathcal{Y}} p(x,y) \log\left(\frac{p(x,y)}{p(y)p(x)}\right) dx dy$$

The mutual information between two random variables measures how much *information* they share.



## 7.2 Kullback-Leibler-Divergence

The *Kullback-Leibler Divergence* is a measure for the difference between two distributions. Even if it is not a real metric, as it is not symmetric, it is a popular tool to compare distributions. For two distributions $p(X)$ and $q(X)$ of continuous random variable $X$, it is defined as:

$$D_{KL}(p(X)||q(X)) = \int_{-\infty}^{\infty} p(x)\log(\frac{p(x)}{q(x)})dx$$

where $p(x)$ and $q(x)$ are the density functions of $p(X)$ and $q(X)$. For discrete probability distributions $p(X)$ and $q(X)$ it is defined as:

$$D_{KL}(p(X)||q(X)) = -\sum_{x \in X} p(x)\log(\frac{p(x)}{q(x)})$$

Derivative of the ELBO:

$$\begin{aligned}
\log(p(x^{(i)})) &= \int_Z p(z|x^{(i)})\log(p(x^{(i)}))dz \\
&= \mathbb{E}_{z \sim p(Z|x^{(i)})}\log(p(x^{(i)})) \\
&= \mathbb{E}_{z \sim p(Z|x^{(i)})}\log\left(\frac{p(x^{(i)}|z)p(z)}{p(z|x^{(i)})}\right) \\
&= \mathbb{E}_{z \sim p(Z|x^{(i)})}\log\left(\frac{p(x^{(i)}|z)p(z)}{p(z|x^{(i)})}\frac{q(z|x^{(i)})}{q(z|x^{(i)})}\right) \\
&= \mathbb{E}_{z \sim p(Z|x^{(i)})}\log(p(x^{(i)}|z)) \\
&\quad + \mathbb{E}_{z \sim p(Z|x^{(i)})}\log\left(\frac{q(z|x^{(i)})}{p(z|x^{(i)})}\right) \\
&\quad + \mathbb{E}_{z \sim p(Z|x^{(i)})}\log\left(\frac{p(z)}{q(z|x^{(i)})}\right) \\
&= \mathbb{E}_{z \sim p(Z|x^{(i)})}\log(p(x^{(i)}|z)) \\
&\quad + KL(q(z|x^{(i)})||p(z|x^{(i)})) \\
&\quad - KL(q(z|x^{(i)})||p(z)) \\
&\geqq \mathbb{E}_{z \sim p(Z|x^{(i)})}\log(p(x^{(i)}|z) \\
&\quad - KL(q(z|x^{(i)})||p(z))
\end{aligned}$$

## 7.3 Generative Adversarial Network

The discriminator $D$ of a GAN should learn a distribution representing the probability that the input is a sample from $p(X)$ compared to $q(X)$. Thus:

$$D(x) = \frac{p(x)}{p(x) + q(x)}$$

Thus, the terms of the minimax game can be written as:

$$\begin{aligned}
& \mathbb{E}_{x \sim p(X)}[\log(D(x))] + \mathbb{E}_{z \sim p(Z)}[\log(1 - D(G(z))] \\
&= \mathbb{E}_{x \sim p(X)}[\log(\tfrac{p(x)}{p(x)+q(x)})] + \mathbb{E}_{z \sim p(Z)}[\log(1 - \tfrac{p(G(z))}{p(G(z))+q(G(z))})] \\
&= \log(\tfrac{1}{2})\mathbb{E}_{x \sim p(X)}[\log(\tfrac{p(x)}{\frac{1}{2}(p(x)+q(x))})] \\
&\quad + \log(\tfrac{1}{2})\mathbb{E}_{z \sim p(Z)}[\log(2 - \tfrac{p(G(z))}{\frac{1}{2}(p(G(z))+q(G(z)))})] \\
&= -\log(4)((\mathbb{E}_{x \sim p(X)}[\log(\tfrac{p(x)}{\frac{1}{2}(p(x)+q(x))})] \\
&\quad \mathbb{E}_{z \sim p(Z)}[\log(\tfrac{q(G(z))}{\frac{1}{2}(p(G(z))+q(G(z)))})]
\end{aligned}$$

Therefore, playing this minimax game is a optimization of the Jensen-Shannon-Divergence:

$$\begin{aligned}
JSD(p(X), q(X)) &= \tfrac{1}{2}D_{KL}(p(X)||\tfrac{1}{2}(p(X)+q(X))) \\
&\quad + \tfrac{1}{2}D_{KL}(q(X)||\tfrac{1}{2}(p(X)+q(X)))
\end{aligned}$$

for two probability distributions $p(X)$ and $q(X)$.

## 7.4 Wasserstein-1 metric, Earth-Mover-Distance

The Wasserstein metric of two distributions $p(X)$ and $q(X)$ of a random variable $X$ is defined as:

$$W(p(X), q(X)) = \inf_{\gamma \in \Pi(p(X), q(X)} \mathbb{E}_{(x,y) \sim \gamma}[\|x - y\|]$$

where $\Pi(p(X), q(X))$ denotes the set of all possible joint distributions of the marginal distributions $p(X)$ and $q(X)$ [Arjovsky et al., 2017].